\documentclass[journal]{IEEEtran}
\usepackage{graphicx}
\usepackage{subcaption}
\usepackage{multirow}
\usepackage[table]{xcolor}
\usepackage{color}
\usepackage{colortbl}
\usepackage{tabularx}
\usepackage{bm}
\usepackage{booktabs}
\usepackage{url}
\usepackage{stackengine}
\usepackage{float}
\usepackage{verbatim}
\usepackage{array}
\usepackage{cite}
\usepackage{algorithm}
\usepackage{algorithmic}
\usepackage[colorlinks,linkcolor=blue]{hyperref}
\usepackage{amsmath,amssymb} 
\usepackage{comment}

\begin{document}
\title{Joint Spatio-Temporal Modeling for Semantic Change Detection in Remote Sensing Images}

\author{Lei~Ding, Jing Zhang, Haitao Guo, Kai Zhang, Bing Liu and Lorenzo~Bruzzone,~\IEEEmembership{Fellow,~IEEE}~

\thanks{L. Ding, H. Guo and B. Liu are with the Information Engineering University, ZhengZhou, China (E-mail: dinglei14@outlook.com, ghtgjp2002@163.com, liubing220524@126.com).}

\thanks{J. Zhang, and L. Bruzzone are with the Department of Information Engineering and Computer Science, University of Trento, 38123 Trento, Italy (E-mail: jing.zhang-1@studenti.unitn.it, lorenzo.bruzzone@unitn.it).}

\thanks{K. Zhang is with School of Information Science and Engineering, Shandong Normal University, Ji'nan 250358, China (E-mail: zhangkainuc@163.com).}

\thanks{This document is funded by the National Natural Science Foundation of China (No. 42201443).}}

\markboth{Manuscript under review}%
{Shell \MakeLowercase{\textit{et al.}}: Bare Demo of IEEEtran.cls for IEEE Journals}

\maketitle

\begin{abstract}
Semantic Change Detection (SCD) refers to the task of simultaneously extracting the changed areas and the semantic categories (before and after the changes) in Remote Sensing Images (RSIs). This is more meaningful than Binary Change Detection (BCD) since it enables detailed change analysis in the observed areas. Previous works established triple-branch Convolutional Neural Network (CNN) architectures as the paradigm for SCD. However, it remains challenging to exploit semantic information with a limited amount of change samples. In this work, we investigate to jointly consider the spatio-temporal dependencies to improve the accuracy of SCD. First, we propose a Semantic Change Transformer (SCanFormer) to explicitly model the 'from-to' semantic transitions between the bi-temporal RSIs. Then, we introduce a semantic learning scheme to leverage the spatio-temporal constraints, which are coherent to the SCD task, to guide the learning of semantic changes. The resulting network (SCanNet) significantly outperforms the baseline method in terms of both detection of critical semantic changes and semantic consistency in the obtained bi-temporal results. It achieves the SOTA accuracy on two benchmark datasets for the SCD.
\end{abstract}




\section{Introduction}\label{sc1}

In Earth observation applications, it is important to locate the areas on the ground affected by semantic changes and to identify what the changes are. Semantic Change Detection (SCD) \cite{bovolo2015time, daudt2019multitask} is the task that addresses these issues by extracting the change areas and the 'from-to' transition information in multi-temporal Remote Sensing (RS) data. This is valuable for a variety of real-world applications, such as Land Cover Land Use (LCLU) monitoring, resource management, disaster alarming, etc.

\begin{figure}[t]
\centering
    \includegraphics[width=1\linewidth]{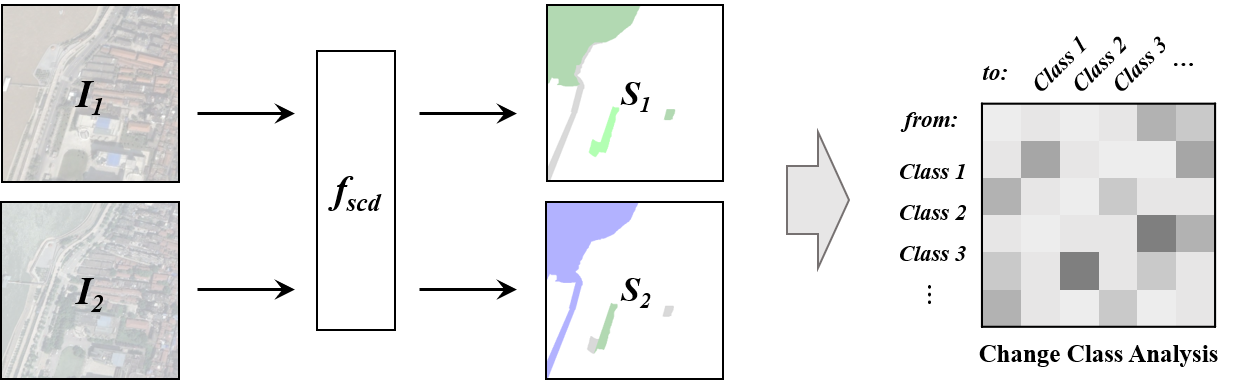}
    \caption{Illustration of the SCD task.}
    \label{Fig.SCD_BCD}
\end{figure}

Differently from the binary change detection (BCD) task that generates only a change map, in SCD it is required to predict the semantic maps before and after the changes. For bi-temporal input images, the results are two semantic change maps associated with each observation time. An illustration of the SCD task is presented in Fig.\ref{Fig.SCD_BCD}. With the results of SCD, it is possible to perform detailed change class analysis in the observed regions by detecting the LCLU transitions.

Previous research reveals that to learn the semantic changes, it is important to model the bi-temporal dependencies. The method in \cite{xian2009updating} first analyzes the change vectors to detect changes, after which classifies the LC classes in changed regions. In the pioneering studies in \cite{bruzzone1997iterative, bruzzone1999neural}, the Bayes rule is applied to iteratively estimate the LCLU transition, which takes into account the prior joint probabilities of pre-change and after-change semantic classes. In \cite{wu2017post} the Bayes rule is also used to fuse the multi-temporal semantic maps with the change probability map to obtain the SCD results.

Recent works employ Convolutional Neural Networks (CNNs) to perform SCD in an end-to-end manner. This avoids the error accumulation problems which are common in the pre-Deep-Learning eras and allows the learned models to generate wider scenes. A widely accepted paradigm for bi-temporal SCD is a triple-branch CNN architecture ~\cite{daudt2019multitask, yang2020asymmetric}. In this architecture, two CNN branches are designed to extract the temporal semantic information, while another CNN branch is deployed to embed the bi-temporal features into the change information. Therefore, the semantic information and change information are separately modeled (instead of being entangled), which fits in with the intrinsic mechanism in the SCD\cite{ding2022bi}.

However, two severe challenges remain to be addressed. The first one is the discrimination of the semantic changes. Apart from salient changes (e.g., emergence of buildings), there are critical changes that are non-salient in local areas (e.g., degeneration of vegetation). Additionally, differences in appearance, illumination and occlusions can be easily confused with semantic changes. Due to these factors, missed alarms and false alarms are common in the results. The second challenge is the discrepancy between the bi-temporal results. In literature works, the bi-temporal semantic correlation is not well-considered, which often causes self-contradictory results (e.g., an area is segmented as \textit{change} but the segmented semantic classes are the same). 

In this work, we investigate the joint modeling of the spatial-temporal dependency to improve the learning of semantic changes. Our major contributions include:

\begin{enumerate}
    \item Proposing a \textit{SCanNet (Semantic Change Network)} to learn the semantic transitions in SCD. The SCanNet first leverages triple 'encoder-decoder' CNNs to learn semantic and change features, then introduces the SCanFormer to learn 'semantic-change' dependencies. The SCanFormer is a variant of the CSWin Transformer, which enables deep and long-range modeling of temporal correlations in the semantic space. Compared to existing methods, the SCanNet shows advantages in discriminating the non-salient semantic changes.
    \item Proposing a \textit{semantic learning scheme} that takes into account the task-specific prior information. The bi-temporal consistency (intrinsic to the SCD task) is utilized as extra supervision to guide the exploitation of semantic information and to reduce the discrepancy in bi-temporal results. We formulate and incorporate two SCD-specific constraints into a learning scheme considering the cases w./w.o. semantic labels in \textit{change}/\textit{no-change} areas, respectively.
\end{enumerate}

The remainder of this paper is organized as follows. In Sec.\ref{sc2} we review the literature works related to BCD, SCD and vision Transformers. In Sec.\ref{sc3} we introduce the proposed method. Sec.\ref{sc4} and Sec.\ref{sc5} report the experimental setups and the obtained results, respectively. Finally, we draw conclusions of this study in Sec.\ref{sc6}.
\section{Related Work}\label{sc2}
This section is organized following the development of CD methods. The pre-CNN and CNN-based methods are separately introduced. Recent methods for the SCD are also reviewed.

\subsection{Binary Change Detection}

In the pre-Deep-Learning era, one possible way to classify change detection methods is to divide them into three categories based on the types of analyzed change features, including texture features, object-based features and angular features \cite{wen2021change}. The CNNs can better capture the context in RSIs, thus they have been widely used for BCD in recent years. Early works on deep-CNN-based CD treats CD as a segmentation task, and employ UNet-like CNNs to directly segment the changes \cite{peng2019end}. In \cite{daudt2018fully} a paradigm for BCD is established, which uses two Siamese CNN encoders (i.e., weight-sharing CNNs) to extract temporal features, and transforms the features into change representations through a CNN decoder.

One of the objectives of CNN-based BCD methods is to discriminate the semantic changes from seasonal changes and spatial misalignment. To extract the change features, feature difference operations are commonly used in the CNNs methods \cite{daudt2018fully, zhang2020feature}. Since there is redundant spatial information in high-resolution (HR) RSIs, multi-scale features are often utilized to filter the spatial noise \cite{hou2021high}. The attention mechanism is also effective to detect semantic changes. In \cite{li2022remote} the channel attention is employed to refine the multi-scale features. In \cite{chen2020dasnet} and \cite{shi2021deeply} hybrid spatial and channel attentions are used to embed discriminative change representations.

\subsection{Semantic Change Detection}

Different from BCD which only detects 'where' the changes are, SCD provides information on 'what' the changes are, which is more valuable in many RS applications. In \cite{singh1990digital} the post-classification comparison method is introduced, which compares the multi-temporal results to classify the LCLU transitions. However, the temporal correlations are not considered in this method. In \cite{bruzzone1997iterative, bruzzone1999neural} the compound classification method is proposed, which applies Bayes rules to iteratively maximize the posterior joint probabilities of multi-temporal LCLU classes. In \cite{wu2017post} the posterior joint probabilities are considered to integrate the change probabilities, which are obtained through slow feature analysis.

The method in \cite{mou2018learning} is an early attempt to apply neural networks for SCD. It is a joint CNN-RNN network where the CNN extracts semantic features, while the RNN models temporal dependencies to classify multi-class changes. In \cite{daudt2018fully} CNN architectures for SCD are elaborately analyzed and a triple-branch CNN architecture is suggested. Two of the branches are deployed to exploit temporal semantics, while another branch is deployed to model the binary change information. In \cite{yang2020asymmetric} a well-annotated benchmark dataset for SCD is released and several evaluation metrics are proposed. In \cite{ding2022bi} a SSCDl (disentangled semantic segmentation and CD, late fusion) architecture is proposed, which has been demonstrated to be more effective for the SCD. 

One of the challenges in the CNN-based methods is the learning of temporal dependencies. While the multi-temporal semantic features are separately embedded, it is important to bridge the connections between different temporal branches. In \cite{ding2022bi} a temporal spatial attention design is proposed to model the bi-temporal semantic correlations. In \cite{peng2021scdnet} the channel attention is used to embed change information into the temporal features. In \cite{zheng2022changemask} a learnable symmetric transform is proposed to align the temporal features and to learn the change maps.

Although these modules to some extent improve the feature representations, the modeling of spatio-temporal dependencies is still partial, one-sided, or shallow. We argue that to learn the intrinsic mechanism in SCD, it is important to jointly model the semantic-change correlations, which is the objective of this study.

\subsection{Vision Transformer}

Vision transformer is a recent emerging research topic since its application in visual recognition tasks \cite{dosovitskiy2020image}. Differently from CNNs whose computations are limited to sliding kernels, vision transformers leverage self-attention to model global dependencies. The vision transformers have been first used in high-level recognition tasks such as image classification \cite{han2021transformer, chu2021twins} and object detection \cite{carion2020end}. Then they have been applied to down-stream tasks such as semantic segmentation\cite{xie2021segformer} and image synthesis \cite{esser2021taming}. Several literature works propose pure-CNN backbone networks for general vision tasks. They arrange hierarchical transformer blocks in CNN-like manner and limit the range of self-attention in local areas to reduce computation\cite{liu2021swin, wang2021pyramid, dong2022cswin}.

There are also investigations to apply transformers to better recognize the complex ground objects in RSIs. In \cite{ding2022looking} a context transformer is proposed to exploit wide-range context information through a context window. In \cite{bandara2022transformer} two siamese transformers are organized into a SegFormer-like \cite{xie2021segformer} architecture for BCD in RSIs. Instead of using transformer as backbone networks feature extraction, the method in \cite{chen2021remote} employs transformer to model the bi-temporal context correlations. In a recent study, transformer is also used in the SCD task \cite{yuan2022transformer}, which is also a variant of the SegFormer. However, the transformers are simply employed as feature extractors to replace the CNN counterparts. Differently, in this work, we investigate leveraging the transformers to deeply model the long-range spatio-temporal correlations that are coherent in the SCD task.
\section{Proposed SCanNet for SCD}\label{sc3}
This section introduces the proposed SCanNet (Semantic Change Network) for SCD in RSIs. Fig.\ref{Fig.Overview} provides an overview of the SCanNet architecture. This is a hybrid framework that consists of triple CNNs and a Transformer. First, temporal features and the change representations are extracted through a Triple Encoder-Decoder (TED) network. Then, we introduce the SCanFormer to learn jointly the spatio-temporal dependencies in the token space. Finally, a semantic learning scheme is proposed to guide the SCanNet toward modeling the intrinsic change-semantic correlations in SCD. Below we introduce the main components of the proposed method.

\subsection{CNN Architecture for SCD} \label{sc3.archs}

Given a pair of input images $(I_1, I_2)$, the task of SCD is to generate a pair of semantic change maps $(Y_1, Y_2)$ that present the changed areas and their semantic categories. An ideal SCD function can be formulated as:
\begin{equation}
    F_{scd}(I_1, I_2) =
    \begin{cases}
      (0, 0), & L_{1}^p = L_{2}^p\\
      (L_1^p, L_2^p), & L_1^p \neq L_2^p
     \end{cases}
     , \forall p \in I
\end{equation}
where $p$ denotes a spatial position on ${I_1, I_2}$ and $L_1^p, L_2^p$ are the bi-temporal semantic classes at $p$.

\begin{figure}[t]
\centering
        \subcaptionbox{}
        {\includegraphics[height=3.0cm]{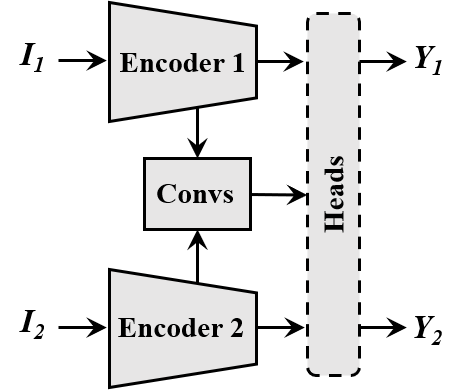}}
        \subcaptionbox{}
        {\includegraphics[height=3.0cm]{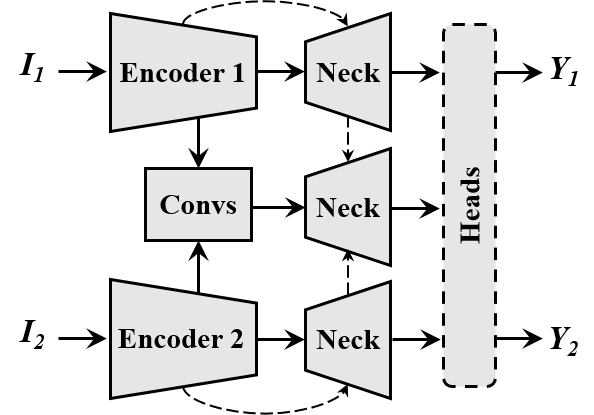}}
    \caption{Comparison of SCD frameworks: (a) SSCDl~\cite{ding2022bi} and (b) the proposed Triple Encoder-Decoder (TED) network. The dash arrows represent skip connections.}\label{Fig.Archs}
\end{figure}

Recent works pointed out that compared to directly learning the semantic changes, separate embedding of the semantic features and change features obtains better accuracy~\cite{daudt2019multitask, yang2020asymmetric, ding2022bi}. In our previous work\cite{ding2022bi}, an SCD framework named SSCDl has been introduced for SCD. As shown in Fig.\ref{Fig.Archs}(a), the SSCDl includes two temporal encoders to exploit the semantic information and a CNN module to detect the changes. There are also optional network modules (presented as 'heads') to enhance the feature representations and to segment the results. One of the remaining problems is that the obtained results are coarse-grained and lack a precise representation of spatial information. Here we improve the SSCDl by adding decoder modules. The resulting TED SCD framework has three feature representation branches organized in an encoder-decoder manner. Fig.\ref{Fig.Archs}(b) presents the TED framework, where the 'Necks' are decoding modules that are attached with lower layers in the encoders.

Let us consider images $I_1, I_2 \in\mathbb{R}^{c\times W\times H}$ input into the TED. First, the encoder networks embed and reshape them into multi-scale features $[X_1^u , X_1^v ], [X_2^u , X_2^v ]$ where $X^u \in\mathbb{R}^{C_u\times W/4\times H/4}, X^v \in\mathbb{R}^{C_v\times W/8\times H/8}$. Then, the necks further enlarge and concatenate $[X_1^u , X_1^v ], [X_2^u , X_2^v ]$ before forwarding them to the heads. Concurrently, there is also a change branch that detects changes by using bi-temporal features. Its inputs are semantic features $[X_1^v , X_2^v ]$ obtained from the highest encoding layers in the temporal branches, while its output is the change feature $X_c^v\in\mathbb{R}^{C_v\times W/8\times H/8}$. The neck in the change branch spatially aligns $X_c$ with temporal features $[X_1^u, X_2^u]$. Finally, outputs of the triple embedding branches are $X_1, X_2, X_c \in \mathbb{R}^{C_v\times W/4\times H/4}$. Compared with the SSCDl, the TED can better retain spatial details.

\subsection{SCanFormer: 'Semantic-Change' dependency modeling with Transformer}

\begin{figure*}[t]
\centering
    \includegraphics[width=1\linewidth]{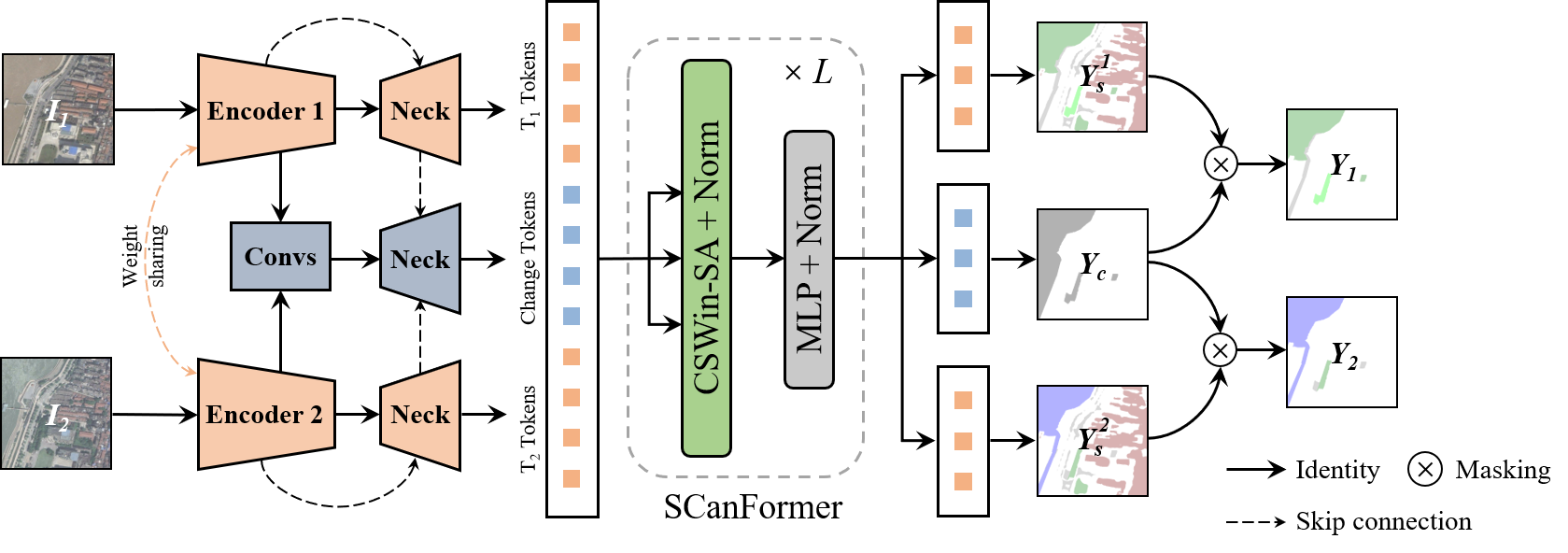}
    \caption{Architecture of the proposed \textbf{ScanNet} (\textbf{S}emantic \textbf{C}h\textbf{an}ge \textbf{Net}work) for SCD.}
    \label{Fig.Overview}
\end{figure*}

In the TED framework, separate embedding of the temporal and change features disentangles the learning of semantic change into Semantic Exploitation (SE) and CD. Although this separate learning enables more dedicated exploitation of the semantic and change information, their correlation is not considered. Intuitively, the bi-temporal semantic differences/consistency helps to discriminate the \textit{changes/no-change} classes, while the LCLU transition pattern allows to better recognize the semantic categories. To model this 'semantic-change' dependency, we resort to the recently emerging transformers and introduce the SCanFormer.

As depicted in Fig.\ref{Fig.Overview}, the SCanFormer can be plugged in the TED framework as the heads. The resulting network, i.e. the SCanNet, is a hybrid 'CNN-Transformer' architecture. The CNN parts serve as feature extractors due to their efficiency and preservation of spatial information, which is the common practice in segmentation tasks \cite{ding2022looking}. Meanwhile, the SCanFormer is attentive to correlations in the embedded semantic space. This allows deep modeling of the 'semantic-change' dependencies in the whole spatio-temporal domain, which differs from previous methods that only model the bi-temporal correlations \cite{ding2022bi} or one-way (change-to-temporal \cite{peng2021scdnet}) attention operations.

To jointly model the spatio-temporal correlations, we concatenate features $[X_1, X_2, X_c]$ and flatten them into a semantic token $\textbf{x} \in \mathbb{R}^{hw\times d}$ where $d$ is the depth of the token (equal to $3C_v$). The SCanFormer consists of $L$ layers of attention blocks. Each of the blocks consists of a Self-Attention (SA) unit and an MLP unit. The units are organized in a residual manner and contain a normalization layer. Mathematically, the calculations inside each attention block are:
\begin{equation}
    \begin{aligned}
    & \hat{\textbf{x}}^{l-1} =  {\rm SA}({\rm LN}(\textbf{x})) + \textbf{x}^{l-1},\\
    & \textbf{x}^{l} =  {\rm MLP}({\rm LN}(\hat{\textbf{x}}^{l-1})) + \hat{\textbf{x}}^{l-1},
\end{aligned}
\end{equation}
where $\textbf{x}^{l}$ represents the output token of the $l$-th block.

The original ViT \cite{dosovitskiy2020image} that uses full SA is calculation-intensive. Its computation complexity is quadratic to the image size, which is expensive for handling HR RSIs. Considering this, we adopt the Cross-Shaped Window (CSWin) SA \cite{dong2022cswin} to model long-range context more efficiently. In the CSWin-SA unit, the input features are spatially partitioned into $M$ vertical and horizontal stripes (in two separate groups), each with the width $s$. There are $2K$ heads to perform SA, $K$ of the heads for the horizontal group and another $K$ ones for the vertical group. The calculations in the horizontal group are:

\begin{equation}
    \begin{aligned}
    & \textbf{x} = [\textbf{x}^1, \textbf{x}^2, ..., \textbf{x}^M],\\
    & \textbf{y}_k^i = A(\textbf{x}^i Q^k, \textbf{x}^i K^k, \textbf{x}^i V^k),\\
    & A_k(\textbf{x}) = [\textbf{y}_k^1, \textbf{y}_k^2, ..., \textbf{y}_k^M]
\end{aligned}
\end{equation}
where $x^i \in\mathbb{R}^{sw\times d}$, $A_k$ represents the SA operation of the $k$-th unit, $Q, K, V \in\mathbb{R}^{d\times d_k}$ are the projecting matrix for the \textit{query, key, value} tokens, respectively. The calculations in the vertical group are similar, thus are omitted for simplicity. The CSWin-SA calculations for the $i$-th stripe are:

\begin{equation}
    \hat{\textbf{x}_i} = [\sigma (\frac{\textbf{q}_i\textbf{k}_i^{T}}{\sqrt{d_k}}) + B_{i} ]\textbf{v}_i,
\end{equation}
where $\sigma$ is the \textit{softmax} operation, $\textbf{q}_i, \textbf{k}_i, \textbf{v}_i$ are linear projections of $\textbf{x}_i$, $B_i$ is a learnable parameter matrix to encode the relative positions. Finally, the outputs of the $2K$ heads are concatenated and projected:

\begin{equation}
    \rm SA(\textbf{x}) = [A_1(\textbf{x}), A_2(\textbf{x}), ..., A_{2K}(\textbf{x})] W^o
\end{equation}
where $W^o\in\mathbb{R}^{d\times C^o}$ is a projection matrix to adjust the token dimension.

\subsection{Semantic Learning with Temporal Consistency Constraints}\label{sc2.loss}

\begin{figure*}[t]
\centering
    \includegraphics[width=1\linewidth]{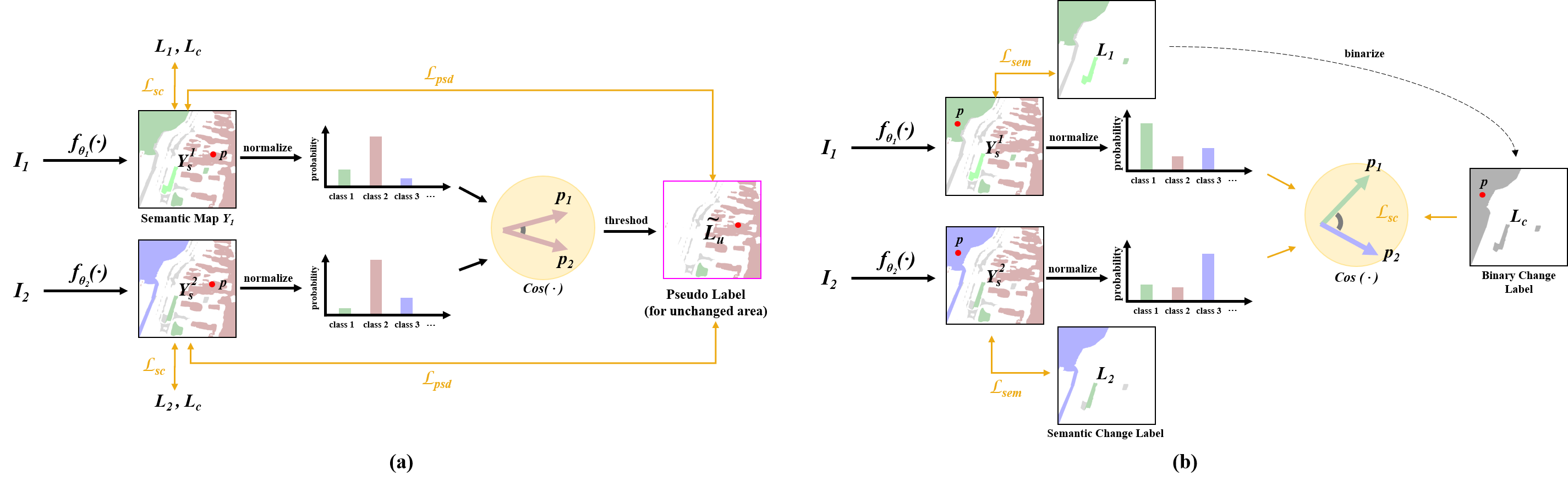}
    \caption{Using temporal consistency as prior constraint to exploit semantic information in (a) \textit{no-change} areas and (b) \textit{change} areas, respectively.}\label{Fig.sem_extract}
\end{figure*}

In the common settings of the SCD task, the changed areas are provided with semantic labels, whereas the unchanged areas are annotated as \textit{no-change}. In other words, the number of semantic labels is very limited. This leads to the challenge of learning semantic information with limited samples. However, considering the intrinsic mechanism in the SCD, it is possible to utilize the bi-temporal consistency as prior information to guide the exploitation of semantic information. Considering the different cases in \textit{change} and \textit{no-change} areas, we propose a semantic learning scheme with two task-specific learning objectives. In the following, we separately discuss the two cases.

First, let us consider the changed areas that are provided with semantic labels. The semantic loss can be calculated with the commonly used cross-entropy loss, defined as:

\begin{equation}
    \mathcal{L}_{sem} = -\sum_{i=1}^2 [L_{i}log(Y_s^i)+(1-L_i)log(1-Y_s^i)]
    \label{L_sem}
\end{equation}
where $i$ is the temporal index. To disentangle SE with CD, this loss is only calculated in the changed areas (i.e. ignoring the \textit{no-change} class).

Meanwhile, the unchanged areas are annotated as \textit{no-change} but usually are not provided with the specific semantic classes. However, it is possible to infer the semantic labels with bi-temporal predictions. Suppose that a place $p$ is known to be unchanged, if its bi-temporal semantic probabilities ${Y_s^1}^p$ and ${Y_s^2}^p$ are similar, we can assume with high confidence that its semantic class should be ${Y_s^1}^p$ (or ${Y_s^2}^p$). Therefore, we adopt the pseudo-labeling method to generate a pseudo semantic label $\widetilde{L}_u$ for $no-change$ areas. Mathematically, the calculation is as follows:

\begin{equation}
    \widetilde{L}_u =\left\{\begin{array}{lr}
                               {Y_s^1}^p, & cos({Y_s^1}^p, {Y_s^2}^p) \geq T \\
                                0, & cos({Y_s^1}^p, {Y_s^2}^p) < T
                            \end{array}, \forall p \in Y_s
                            \right.
\end{equation}
where $cos()$ is a \textit{Cosine} function calculated in the vector space to measure the semantic similarity, and $T$ is a threshold. We use annotation \textit{0} to exclude the change areas from the loss calculation. A pseudo semantic objective $\mathcal{L}_{psd}$ can be calculated as:
\begin{equation}
    \mathcal{L}_{psd} = -\sum_{i=1}^2 [\widetilde{L}_u log(Y_s^i)+(1-\widetilde{L}_u)log(1-Y_s^i)]
    \label{L_psd}
\end{equation}
The calculation of this objective function is illustrated in Fig.\ref{Fig.sem_extract}(a).

Last but not least, according to the intrinsic logic in the SCD, there should be semantic consistency in the \textit{no-change} areas on $Y_1$ and $Y_2$, whereas there is a difference in the changed areas. Using this temporal constraint as prior information, a semantic consistency learning objective $\mathcal{L}_{sc}$ can be constructed to guide the network training. $\mathcal{L}_{sc}$ is calculated with the ground truth change label $L_{c}$, which can be easily derived by binarizing $L_1$ or $L_2$ (setting the annotation of changed regions to $0$). The calculation of$\mathcal{L}_{sc}$ is as follow:
\begin{equation}
    \mathcal{L}_{sc}  =\left\{\begin{array}{lr}
                                1-cos({Y_s^1}^p, {Y_s^2}^p), & L_{c}^p=1  \\
                                cos({Y_s^1}^p, {Y_s^2}^p), & L_{c}^p=0
                            \end{array}, \forall p \in Y_s
                      \right.
\end{equation}
The calculation of this learning objective function is depicted in Fig.\ref{Fig.sem_extract}(b). Note that this function is calculated in both \textit{no-change} and \textit{change} areas. It encourages the network to generate the same semantic predictions in \textit{change} areas, whereas generating different predictions in \textit{no-change} areas.

Incorporating these learning objective functions, the overall loss $\mathcal{L}$ is given by:
\begin{equation}
    \mathcal{L} = \mathcal{L}_{sem} + \mathcal{L}_{psd} + \mathcal{L}_{sc}
\end{equation}
By adding $\mathcal{L}_{sc}$, the temporal semantic information contained in the two images is jointly considered, which improves the discrimination of critical areas.

\subsection{Implementation Details}

In the following, we report the parameter settings and the training details in the implementation of SCanNet.

1) \textbf{TED}. The CNN encoders in TED are constructed following the practice in \cite{ding2022bi}. We adopt the ResNet blocks which are commonly used in segmentation tasks \cite{ding2020lanet}. Since the pre and after-change input images are in the same domain, the temporal encoders are made Siamese (i.e., weight-sharing) to avoid over-fitting and to better align the extracted features. The convolutional layer to embed change features is a ResNet-like block with 6 residual layers. The necks are decoder blocks constructed as those in \cite{chen2018deeplabv3+}.

2) \textbf{ScanFormer}. The important parameters in the SCanFormer are $s$, $L$, and $K$. $s$ determines the context modeling range, which is set to 2 considering the input feature size ($H/4 \times W/4$). To avoid heavy computations, $L$ (the number of layers in the SA blocks) is set to 2. $K$ is set to 2 considering the feature dimensions.

3) \textbf{Training Settings}. The proposed methods are implemented with the PyTorch toolbox. The training is performed for 50 epochs. The batch size and initial learning rate are 8 and 0.1, respectively. The gradient descent optimization method is Stochastic Gradient Descent with Nesterov momentum. The learning rate $lr$ is updated at each iteration as: $0.1*(1-iterations/total\_iterations)^{1.5}$. The augmentation operations consist of random flipping and rotation before loading the input images.

For more details, readers are encouraged to visit our codes released at: \url{https://github.com/ggsDing/SCanNet}.
\section{Dataset Description and Experimental Settings}\label{sc4}
In this section, we describe the dataset, the evaluation metrics and the experimental settings.

\subsection{Dataset}

We conduct experiments on two openly-accessible benchmark datasets for the SCD: the SECOND and the Landsat-SCD datasets. The former is an HR dataset collected in city regions, whereas the latter is a mid-resolution dataset at the margin of a desert area. These differences in the spatial resolution and the observed scenes make it possible to assess the effectiveness of the tested methods in different scenarios.

1) \textbf{SECOND}\cite{yang2020asymmetric}.
The SEmantic Change detectiON Dataset (SECOND) is a large-scale and well-annotated benchmark dataset for the SCD in HR RSIs. This dataset includes 4662 pairs of RSIs acquired by different sensors and platforms. They are obtained in various cities including Hangzhou, Chengdu, and Shanghai in China. Each image has $512 \times 512$ pixels. The ground sampling distance (GSD) in this dataset is in the range between 0.5 and 3m.  
Each pair of RSIs are spatially matched, recording the changes in the considered region. The manual annotations are the semantic change maps associated with each image. In the annotated labels, there are 1 \textit{no-change} class and 5 LC classes, including \textit{non-vegetated ground surface, tree, low vegetation, water, buildings} and \textit{playgrounds} (only the changed areas are annotated). Comparing the pre-change and after-change classes, a total of 30 types of semantic changes can be derived. The change pixels account for 19.87\% of the total image pixels. We further split 1/5 of the data (593 pairs of RSIs) as the test set and use the remaining ones for training (2375 pairs) following the practice in \cite{ding2022bi}.

2) \textbf{Landsat-SCD Dataset}\cite{yuan2022transformer}.
This dataset is constructed with Landsat images collected between 1990 and 2020. The observed region is in Tumushuke, Xinjiang, China, which is at the margin of the Taklimakan Desert. The ground sampling distance is 30m. Changed pixels account for 18.89\% of the total pixels. The dataset is annotated into the \textit{no-change} class and 4 LC classes, including \textit{farmland, desert, building} and \textit{water} (only the changed areas are annotated). These LC classes are associated with 10 types of semantic changes.

The Landsat-SCD dataset consists of 8468 pairs of images, each having the spatial resolution of $416 \times 416$. Excluding those samples obtained by spatial augmentation operations (including flipping, masking and resizing), there are 2425 pairs of original data. We further split them into training, validation and test sets with the numeric ratio of 3:1:1 (i.e., 1455, 485 and 485, respectively).

\subsection{Evaluation Metrics}

To evaluate the segmentation accuracy, we adopt 4 metrics commonly used in the SCD task, including overall accuracy (OA), mean Intersection over Union (mIoU), Separated Kappa (SeK) coefficient and SCD-targeted F1 Score ($F_{scd}$). OA is a common metric in segmentation tasks\cite{ding2021adversarial}, which represents the numeric ratio between the correctly classified pixels and the total image pixels. Let $Q=\{q_{i,j}\}$ be the confusion matrix, where $q_{i,j}$ represents the number of pixels that are classified into class $i$ while their GT index is $j$ ($i, j \in \{0, 1, ..., N\}$ ($0$ represents \textit{no-change}). OA is calculated as:
\begin{equation}
    OA = \sum_{i=0}^{N} q_{ii} / \sum_{i=0}^{N}\sum_{j=0}^{N} q_{ij}.
\end{equation}
In the SCD task, since \textit{no-change} class is the majority, OA cannot well-describe the discrimination of semantic classes. In \cite{yang2020asymmetric} mIoU and SeK are introduced to evaluate the performance CD and SE, respectively. The mIoU in SCD is calculated based on the discrimination of \textit{change}/\textit{no-change} classes:
\begin{gather}
    mIoU = (IoU_{nc} + IoU_{c})/2,\\
    IoU_{nc} = q_{00}/ (\sum_{i=0}^{N} q_{i0} + \sum_{j=0}^{N} q_{0j} - q_{00}),\\
    IoU_{c} = \sum_{i=1}^{N} \sum_{j=1}^{N} q_{ij} / (\sum_{i=0}^{N} \sum_{j=0}^{N} q_{ij} - q_{00}),
\end{gather}

Meanwhile, SeK evaluates the segmentation of semantic classes especially in the \textit{changed} areas. It is calculated based on the confusion matrix $\hat{Q} = \{\hat{q}_{ij}\}$, where $\hat{q}_{ij} = q_{ij}$ except that $\hat{q}_{00} = 0$. This excludes the true positive non-changed pixels whose number is dominant. It is calculated as:
\begin{gather}
    \rho = \sum_{i=0}^{N}\hat{q}_{ii} / \sum_{i=0}^{N}\sum_{j=0}^{N}\hat{q}_{ij},\\
    \eta = \sum_{i=0}^{N} (\sum_{j=0}^{N}\hat{q}_{ij} * \sum_{j=0}^{N}\hat{q}_{ji}) / (\sum_{i=0}^{N}\sum_{j=0}^{N}\hat{q}_{ij}) ^2,\\
    SeK = e^{IoU_c-1} \cdot (\rho - \eta) / (1 - \eta).
\end{gather}

In \cite{ding2022bi} the $F_{scd}$ is introduced to evaluate the segmentation accuracy in \textit{change} areas. It is calculated deriving the  $Precision$ ($P_{scd}$) and $Recall$ ($R_{scd}$) in the areas annotated as \textit{change}:
\begin{gather}
    P_{scd} = \sum_{i=1}^{N} q_{ii} / \sum_{i=1}^{N}\sum_{j=0}^{N} q_{ij},\\
    R_{scd} = \sum_{i=1}^{N} q_{ii} / \sum_{i=0}^{N}\sum_{j=1}^{N} q_{ij},\\
    F_{scd} = \frac{2*P_{scd}*R_{scd}}{P_{scd}+R_{scd}}
\end{gather}


\section{Experimental Results}\label{sc5}
In this section, we present the experimental results obtained on the benchmark datasets. First, ablation studies are conducted to demonstrate the effects of the different components of the proposed techniques. Then, we present the qualitative results obtained on sample test data. Finally, the proposed SCanNet is compared with SOTA methods in SCD.

\begin{table*}[htbp]
\centering
    \caption{Quantitative results of the ablation study related to the proposed techniques.}
    \resizebox{1\linewidth}{!}{%
        \begin{tabular}{l|ccc|cccc}
        \toprule
            \multirow{2}*{Methods} & \multicolumn{3}{c|}{Proposed Techniques}  & \multicolumn{4}{c}{Accuracy}\\
            \cline{2-8}
            & SCanFormer & $\mathcal{L}_{psd}$ & $\mathcal{L}_{sc}$ & OA(\%) & mIoU(\%) & Sek(\%) & $F_{scd}$(\%) \\
            \hline
            SSCDl \cite{ding2022bi} &  &  &  & 87.19 & 72.60 & 21.86 & 61.22  \\
            \hline
            TED &  &  &  & 87.39 & 72.79 & 22.17 & 61.56 \\
            TED (w. $\mathcal{L}_{sc}$)  &  &  & $\surd$ & 87.67 & 73.10 & 22.83 & 62.29 \\
            TED (w. $\mathcal{L}_{sc} \& \mathcal{L}_{psd}$) &  & $\surd$ & $\surd$ & 87.75 & \textbf{73.45} & 23.46 & 62.76 \\
            \hline
            SCanNet (w/o. $\mathcal{L}_{psd}$) &  $\surd$ &  & $\surd$ & 87.63 & 73.09 & 23.42 & 63.10 \\
            SCanNet & $\surd$ & $\surd$ & $\surd$ & \textbf{87.86} & 73.42 & \textbf{23.94} & \textbf{63.66} \\
        \bottomrule
        \end{tabular} }\label{Table.Ablation}
\end{table*}

\subsection{Ablation Study}
1) \textbf{Quantitative Results}. The techniques proposed in this work include i) the TED framework; ii) the semantic learning scheme and iii) the SCanNet (incorporating SCanFormer into the TED). Table.\ref{Table.Ablation} reports the quantitative results after using the proposed techniques. The TED framework outperforms the SSCDl framework by around 0.3\% in mIoU and $F_{scd}$, which demonstrates its advantages in both SE and CD. After applying the semantic learning scheme to the TED, there are significant improvements in all the metrics (around 0.7\% in mIoU, SeK and $F_{scd}$). Compared to the plain TED, the SCanNet (adding the SCanFormer) has significant improvements in SE (demonstrated by improvements of around 0.5\% in Sek and 0.9\% in $F_{scd}$).

\begin{figure}[t]
\centering
    {\includegraphics[width=8.5cm]{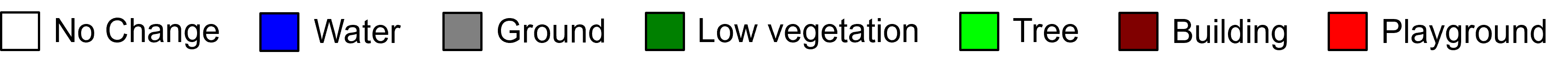}}\\
    \setlength{\tabcolsep}{1pt}
    \begin{tabular}{>{\centering\arraybackslash}m{0.5cm}>{\centering\arraybackslash}m{2.0cm}>{\centering\arraybackslash}m{2.0cm}>{\centering\arraybackslash}m{2.0cm}>{\centering\arraybackslash}m{2.0cm}}
        (a1) &
        \includegraphics[width=1.8cm]{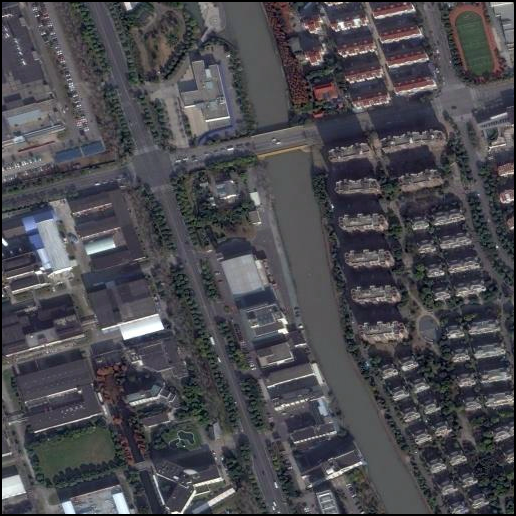} &
        \includegraphics[width=1.8cm]{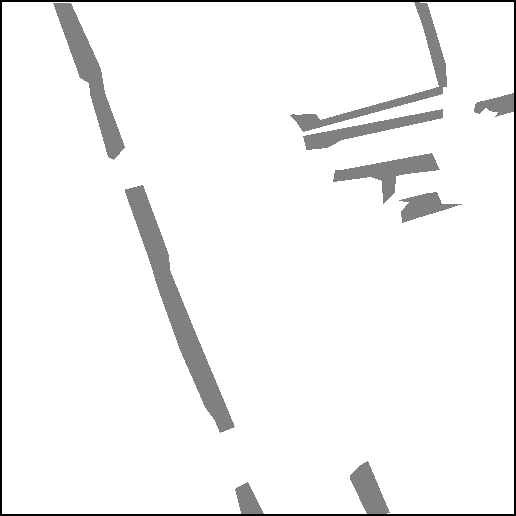} &
        \includegraphics[width=1.8cm]{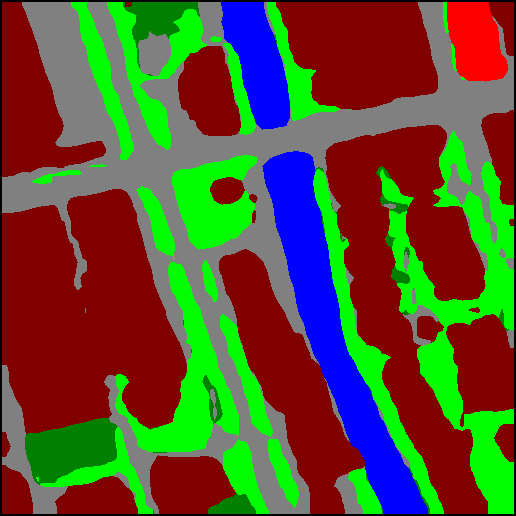} &
        \multirow{2}*{\includegraphics[width=1.8cm]{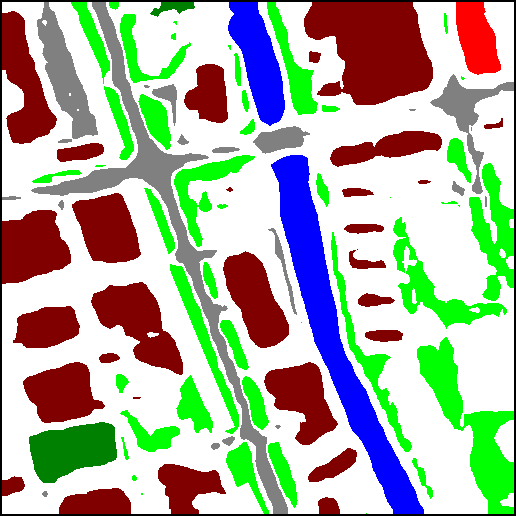}}\\
        (a2) &
        \includegraphics[width=1.8cm]{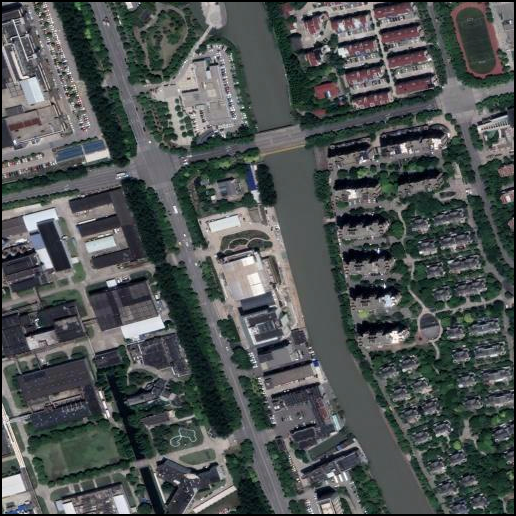} &
        \includegraphics[width=1.8cm]{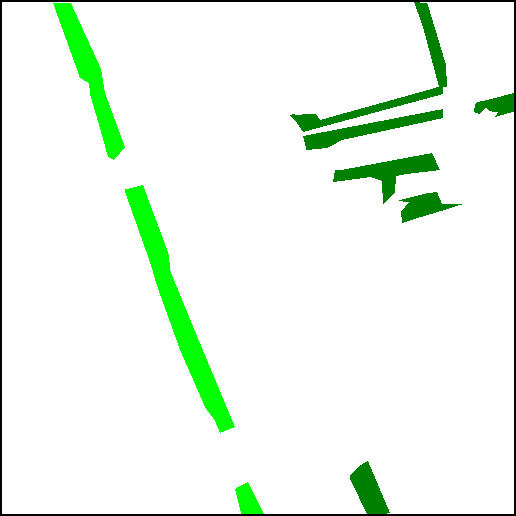} &
        \includegraphics[width=1.8cm]{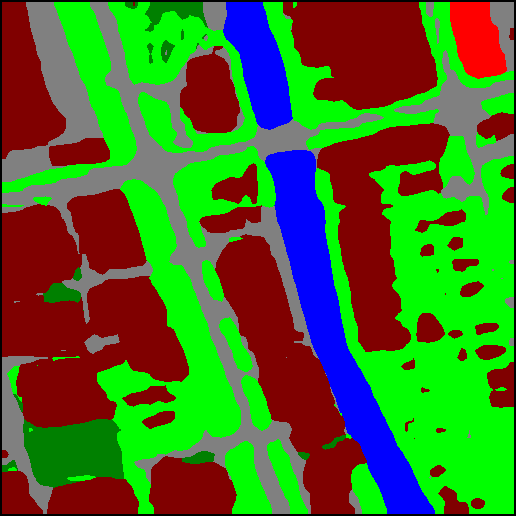} & \\
        \hline\\
        (b1) &
        \includegraphics[width=1.8cm]{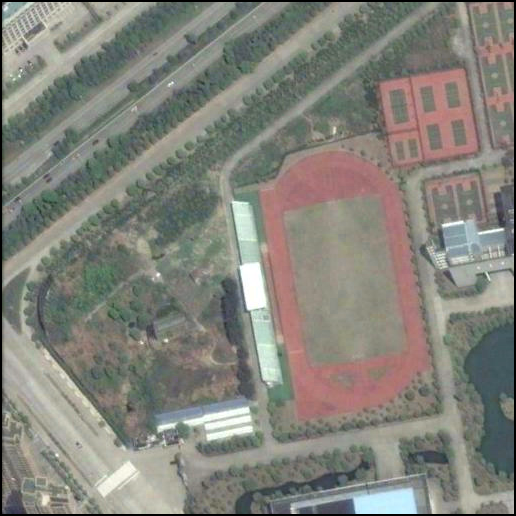} &
        \includegraphics[width=1.8cm]{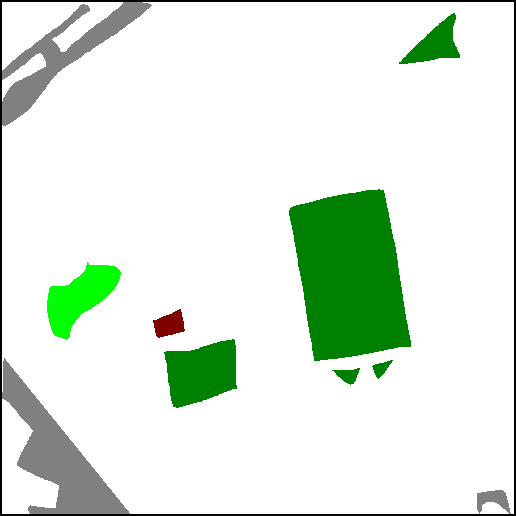} &
        \includegraphics[width=1.8cm]{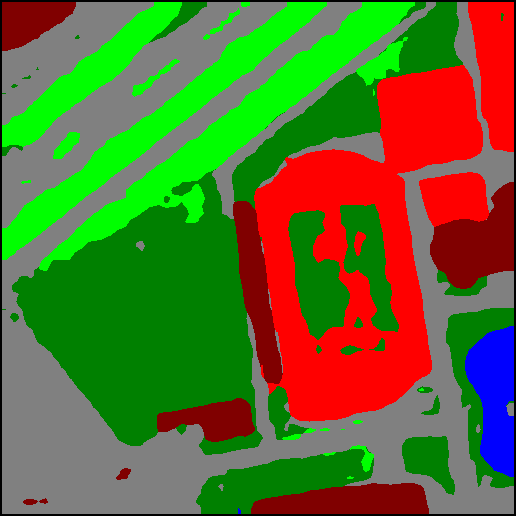} &
        \multirow{2}*{\includegraphics[width=1.8cm]{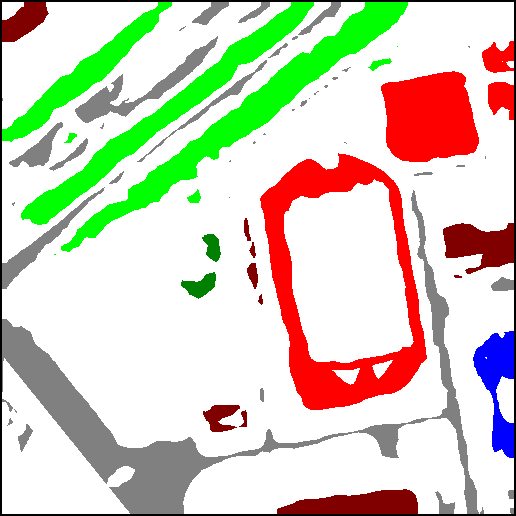}}\\
        (b2) &
        \includegraphics[width=1.8cm]{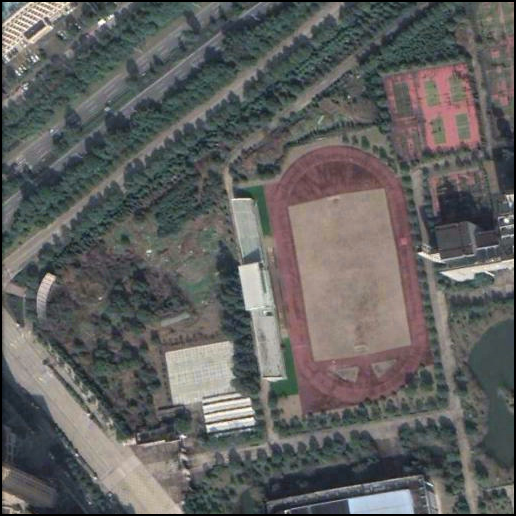} &
        \includegraphics[width=1.8cm]{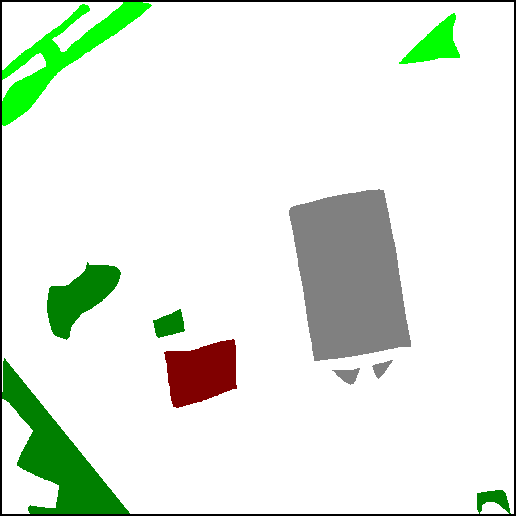} &
        \includegraphics[width=1.8cm]{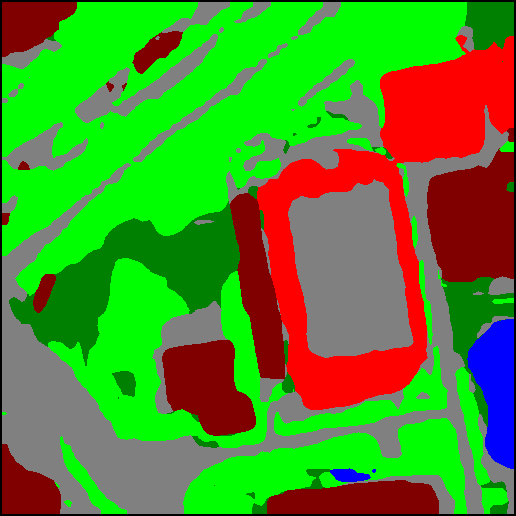} & \\
        & Training & \multirow{2}*{GT} & \multicolumn{2}{c}{$\underbrace{\rm Semantic\:map \; Pseudo\:label}$} \\
        & images & & \multicolumn{2}{c}{(generated by the TED)} \\
    \end{tabular}
    \caption{Generation of the pseudo labels.} \label{Fig.Psd}
\end{figure}

2) \textbf{Effects of the Semantic Learning Scheme}. The $\mathcal{L}_{psd}$ in the proposed semantic learning scheme is calculated with pseudo labels generated by bi-temporal semantic predictions. To visually assess the quality of generated pseudo labels, we present some examples in Fig.\ref{Fig.Psd}. The pseudo labels cover the pixels with no semantic labels, thus their correctness can only be visually assessed together with the input images. One can observe that the generated labels are generally correct and multiple semantic categories are included. This supervision function improves the learning of semantic information in unchanged regions where the bi-temporal prediction confidence is high.

\begin{figure}[t]
\centering
    {\includegraphics[width=8.5cm]{Pics/ST_colorbar.png}}\\
    \setlength{\tabcolsep}{1pt}
    \begin{tabular}{>{\centering\arraybackslash}m{0.6cm}>{\centering\arraybackslash}m{1.5cm}>{\centering\arraybackslash}m{1.5cm}>{\centering\arraybackslash}m{1.5cm}>{\centering\arraybackslash}m{1.5cm}>{\centering\arraybackslash}m{1.5cm}}
        (a1) &
        \includegraphics[width=1.5cm]{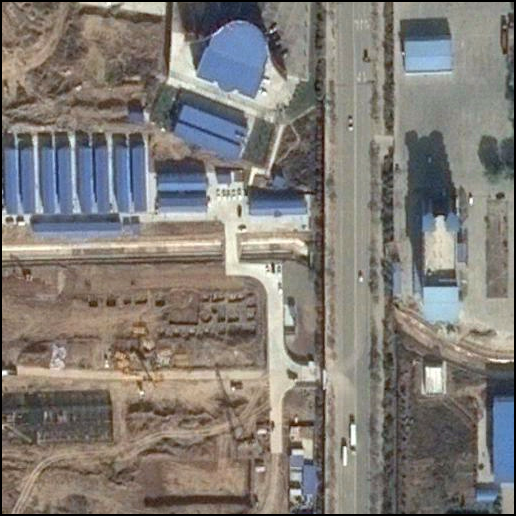} &
        \includegraphics[width=1.5cm]{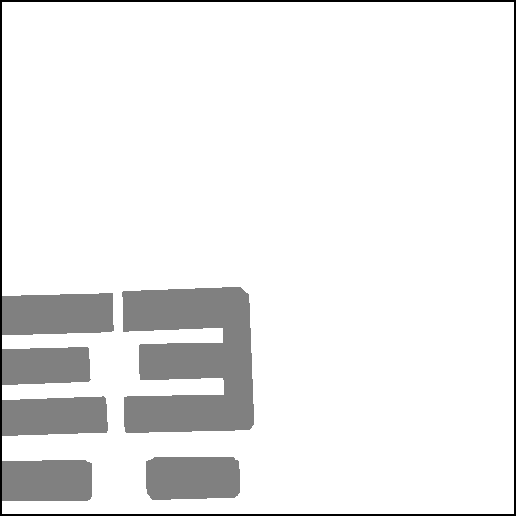} &
        \includegraphics[width=1.5cm]{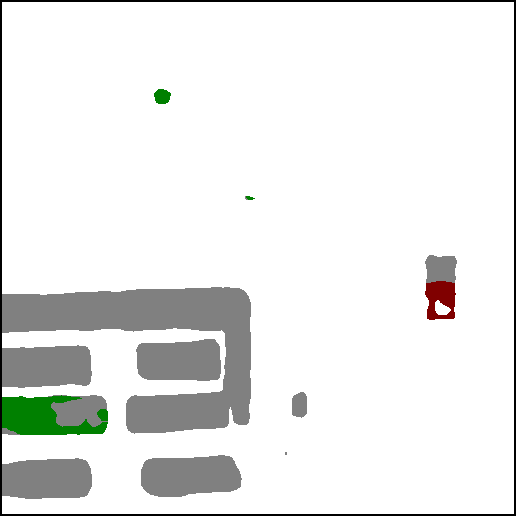} &
        \includegraphics[width=1.5cm]{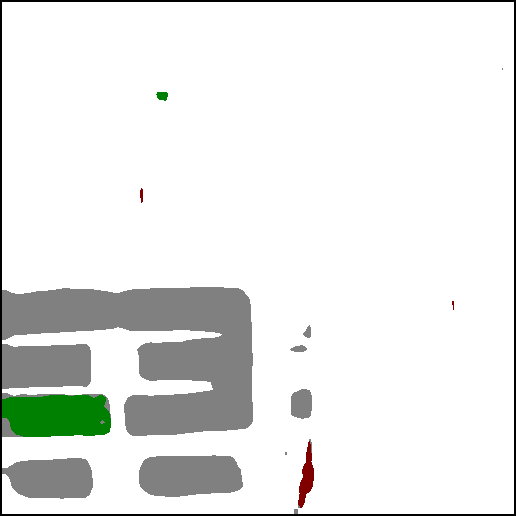} &
        \includegraphics[width=1.5cm]{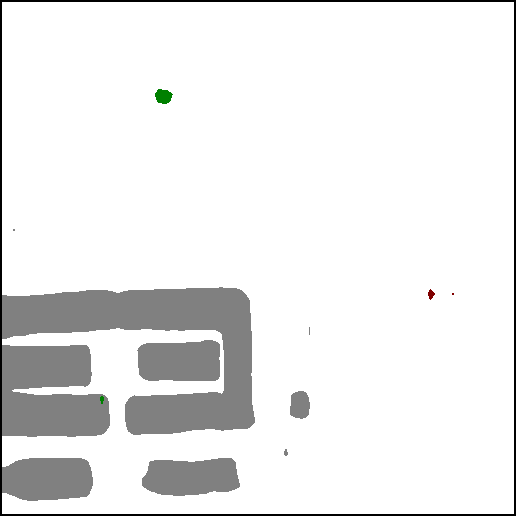} \\
        (a2) &
        \includegraphics[width=1.5cm]{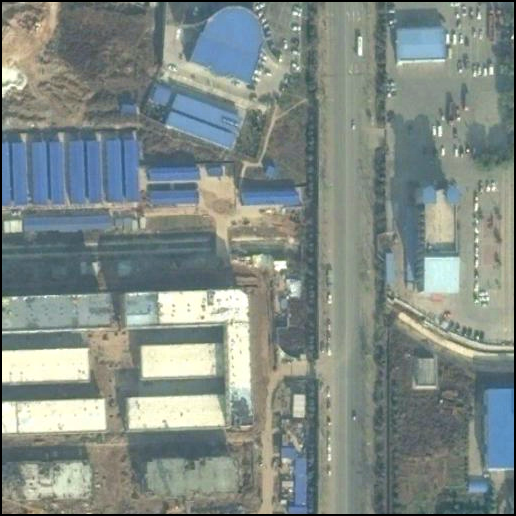} &
        \includegraphics[width=1.5cm]{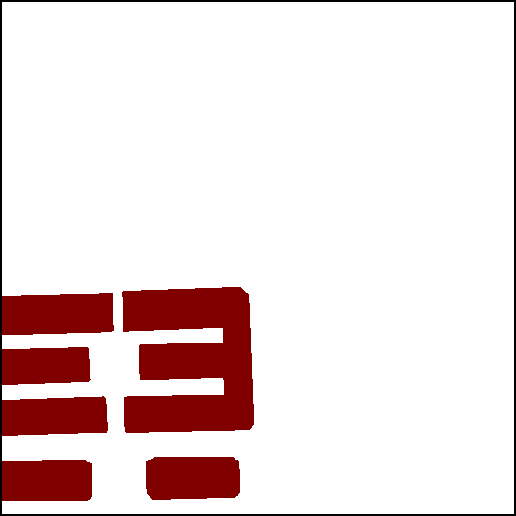} &
        \includegraphics[width=1.5cm]{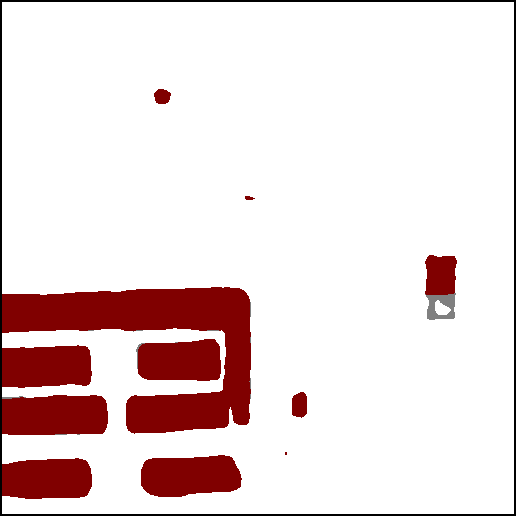} &
        \includegraphics[width=1.5cm]{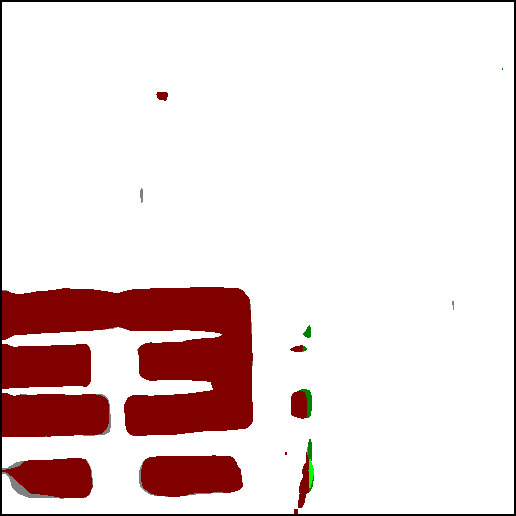} &
        \includegraphics[width=1.5cm]{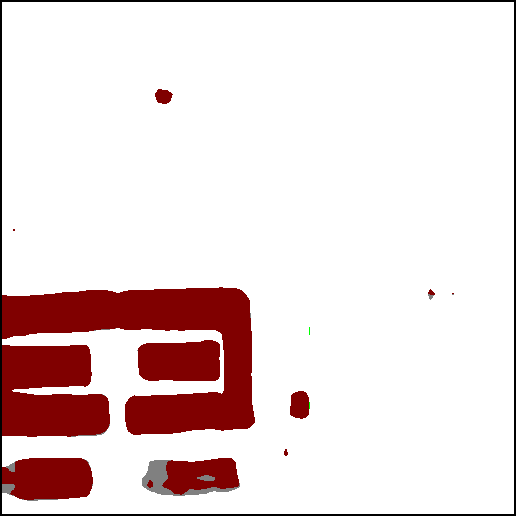} \\
        \hline\\
        (b1) &
        \includegraphics[width=1.5cm]{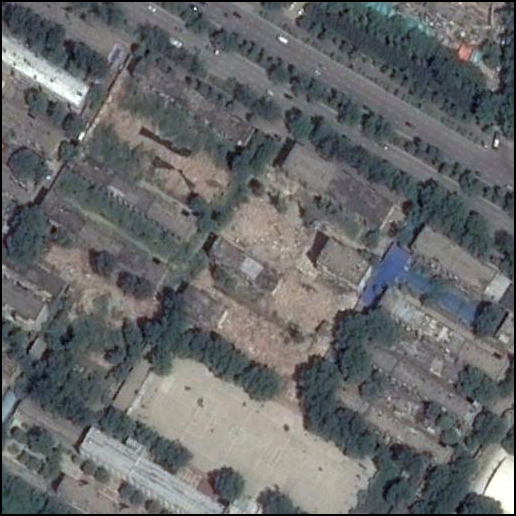} &
        \includegraphics[width=1.5cm]{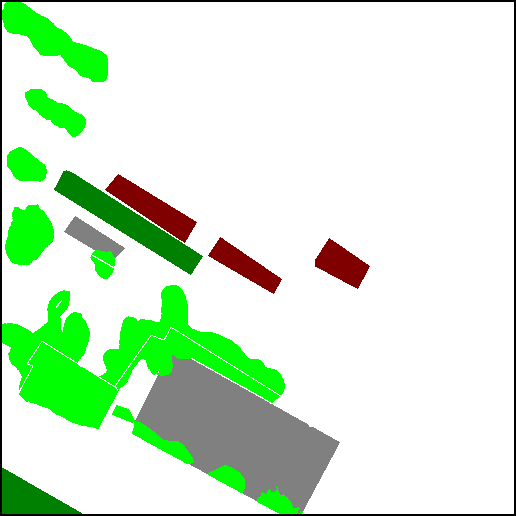} &
        \includegraphics[width=1.5cm]{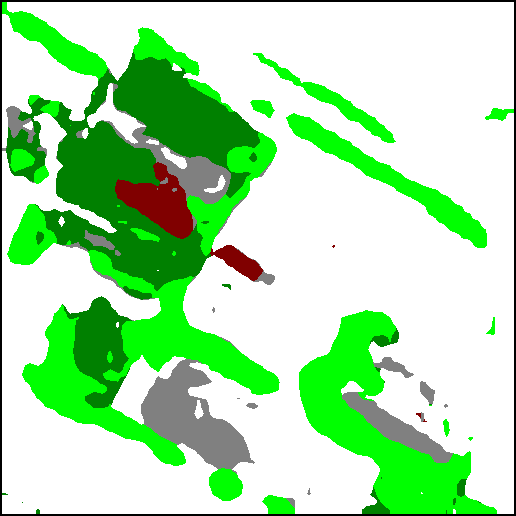} &
        \includegraphics[width=1.5cm]{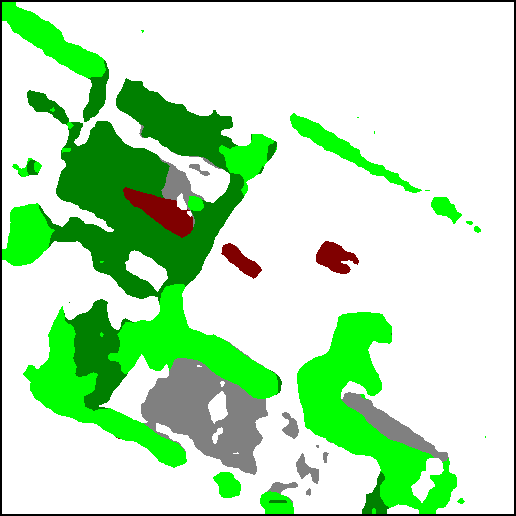} &
        \includegraphics[width=1.5cm]{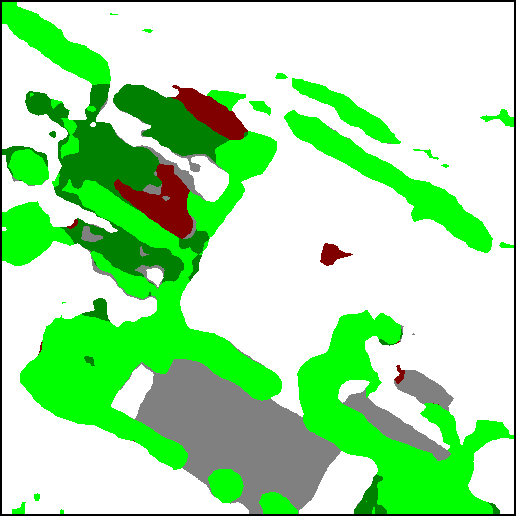} \\
        (b2) &
        \includegraphics[width=1.5cm]{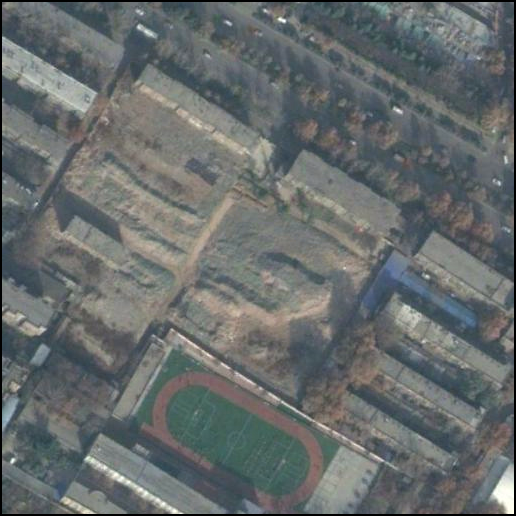} &
        \includegraphics[width=1.5cm]{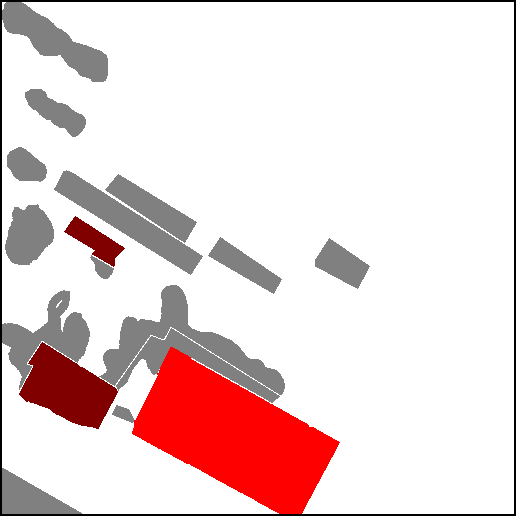} &
        \includegraphics[width=1.5cm]{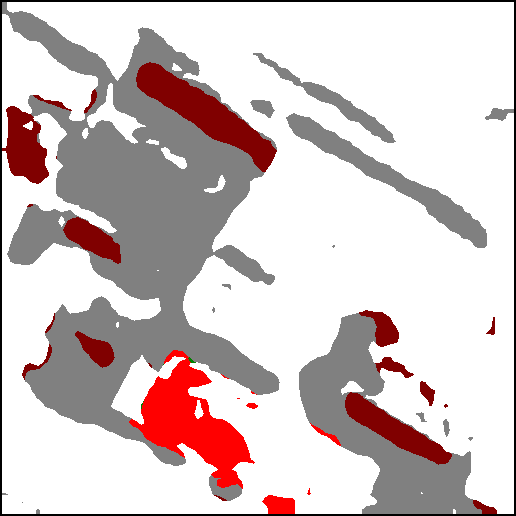} &
        \includegraphics[width=1.5cm]{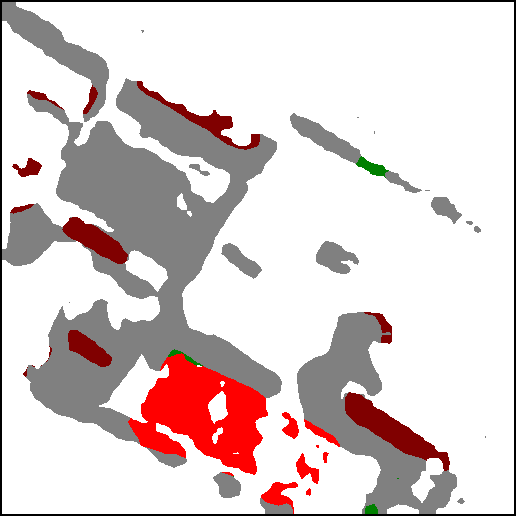} &
        \includegraphics[width=1.5cm]{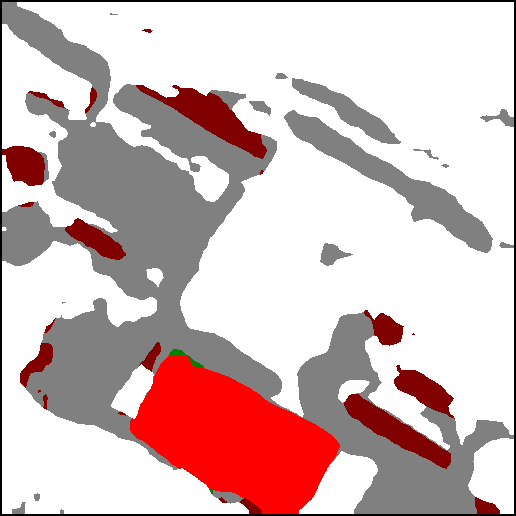} \\
        & Test image & GT & TED & TED (w. $\mathcal{L}_{sc}$) & TED (w. $\mathcal{L}_{sc} \& \mathcal{L}_{psd}$) \\
    \end{tabular}
    \caption{Comparison of results obtained by the TED w. and w/o. using the semantic learning scheme.} \label{Fig.Ablation_psd}
\end{figure}

Fig.\ref{Fig.Ablation_psd} presents qualitative comparisons of the results obtained w. and w./o using the semantic learning scheme. The comparisons are made on the plain TED framework where there is no other explicit spatio-temporal modeling design, so that all the differences can be attributed to the proposed learning scheme. After using the semantic learning scheme, some non-salient changes are captured (e.g., the emergence of a \textit{playground} is detected in Fig.\ref{Fig.Ablation_psd}(b)), while the discrimination of semantic classes is also improved (e.g., the error of segmenting \textit{ground} as \textit{low vegetation} in Fig.\ref{Fig.Ablation_psd}(a)). These results indicate that the proposed learning scheme brings overall improvements to the SCD task.

\begin{figure}[t]
\centering
    {\includegraphics[width=8.5cm]{Pics/ST_colorbar.png}}\\
    \setlength{\tabcolsep}{1pt}
    \begin{tabular}{>{\centering\arraybackslash}m{0.8cm}>{\centering\arraybackslash}m{1.6cm}>{\centering\arraybackslash}m{1.6cm}>{\centering\arraybackslash}m{1.6cm}>{\centering\arraybackslash}m{1.6cm}}
        (a1) &
        \includegraphics[width=1.6cm]{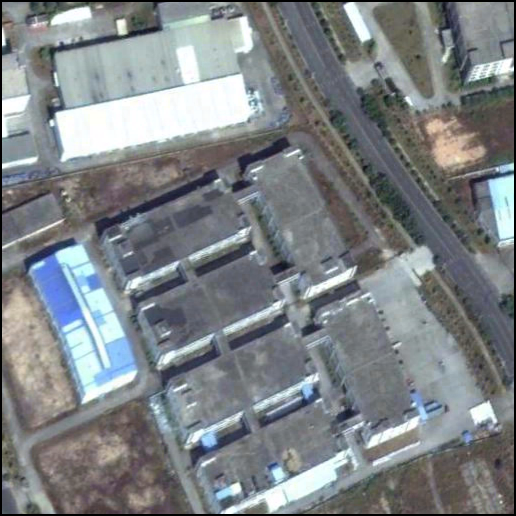} &
        \includegraphics[width=1.6cm]{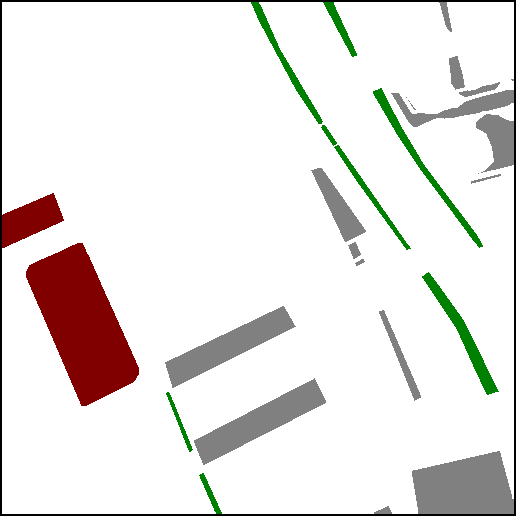} &
        \includegraphics[width=1.6cm]{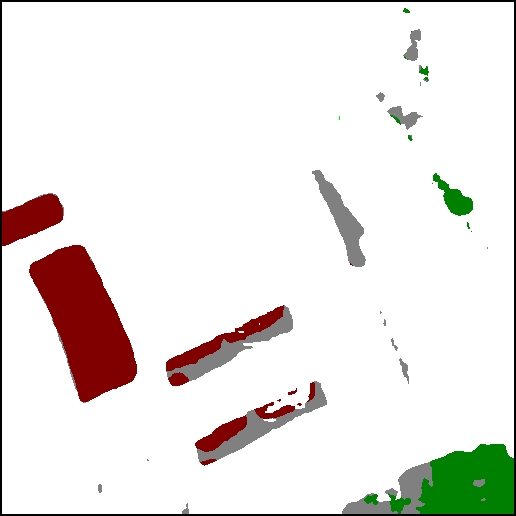} &
        \includegraphics[width=1.6cm]{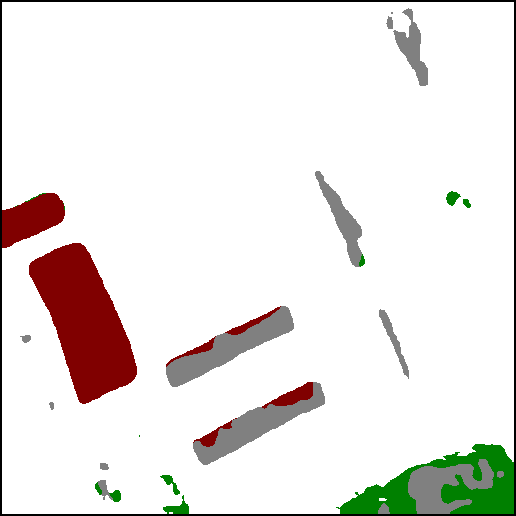} \\
        (a2) &
        \includegraphics[width=1.6cm]{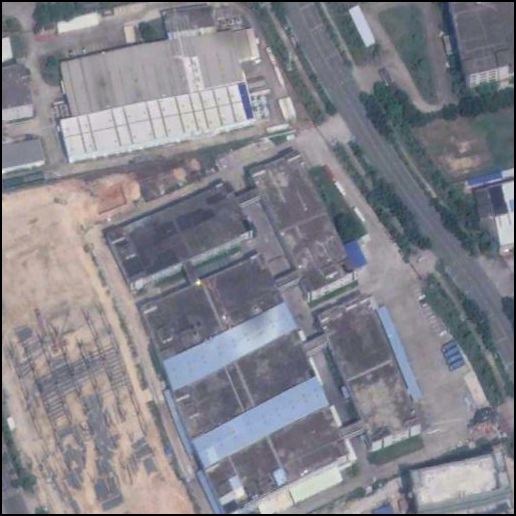} &
        \includegraphics[width=1.6cm]{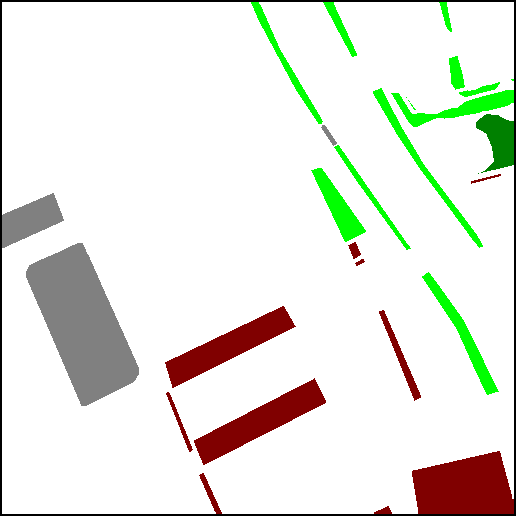} &
        \includegraphics[width=1.6cm]{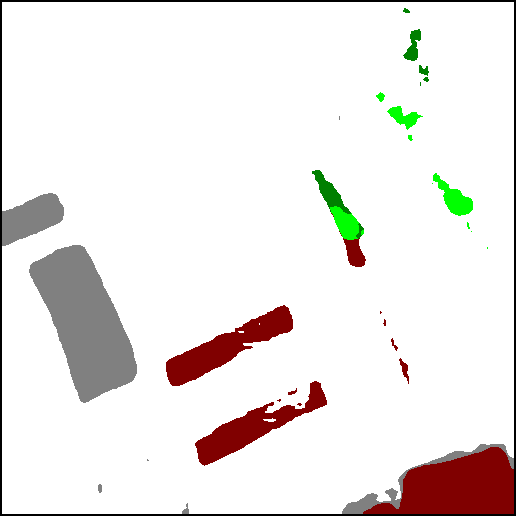} &
        \includegraphics[width=1.6cm]{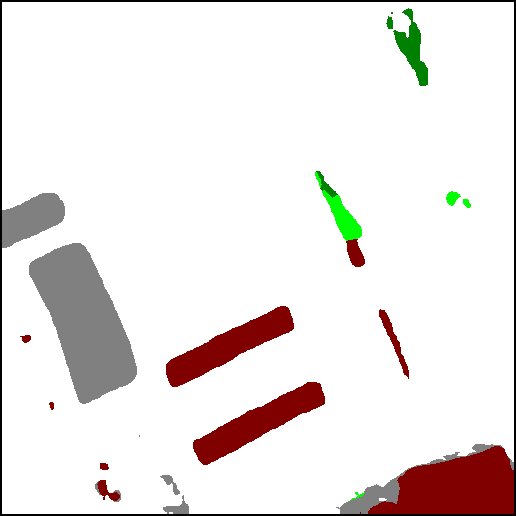} \\
        \hline\\
        (b1) &
        \includegraphics[width=1.6cm]{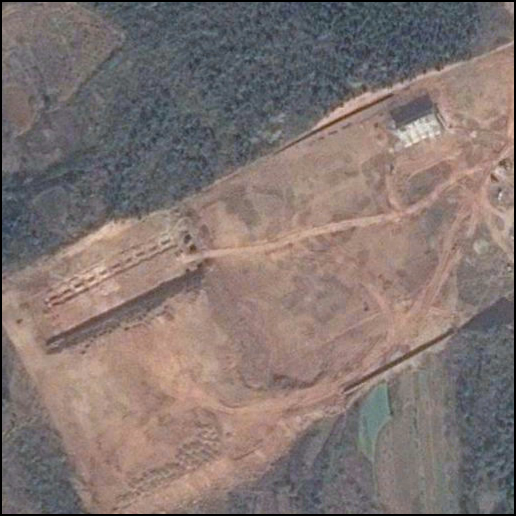} &
        \includegraphics[width=1.6cm]{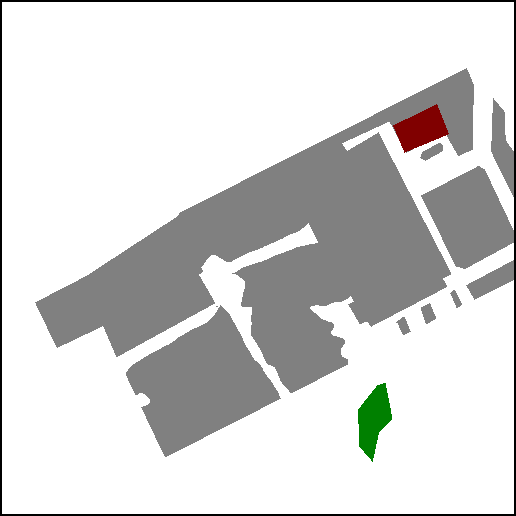} &
        \includegraphics[width=1.6cm]{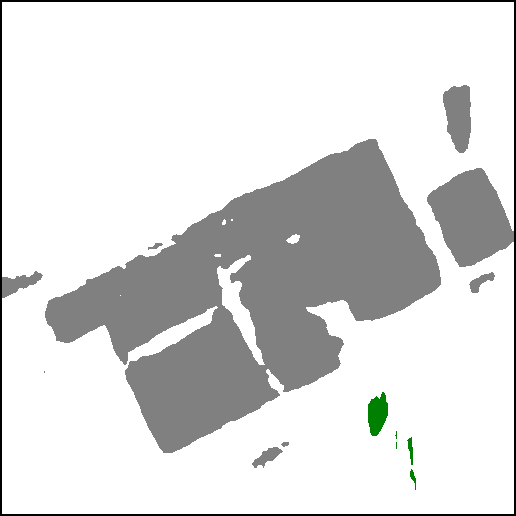} &
        \includegraphics[width=1.6cm]{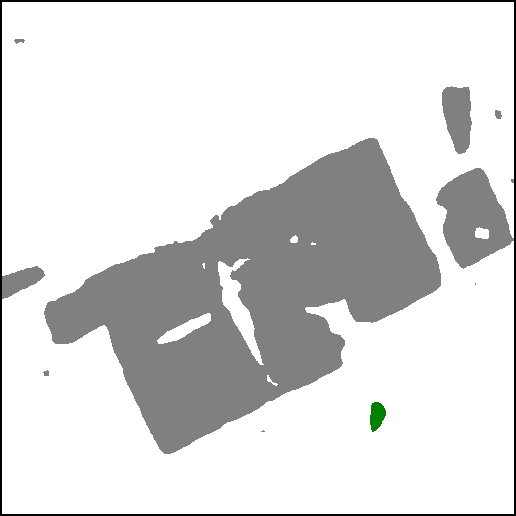} \\
        (b2) &
        \includegraphics[width=1.6cm]{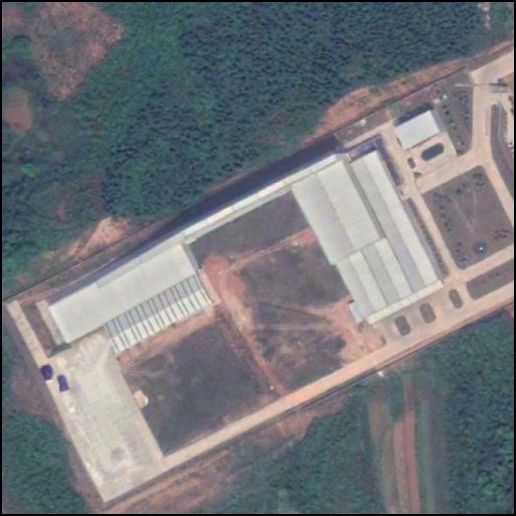} &
        \includegraphics[width=1.6cm]{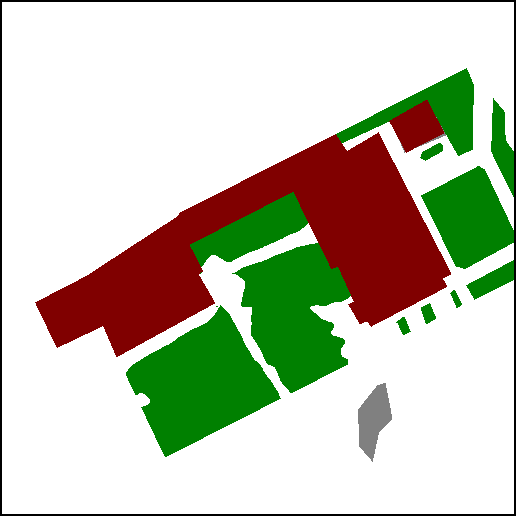} &
        \includegraphics[width=1.6cm]{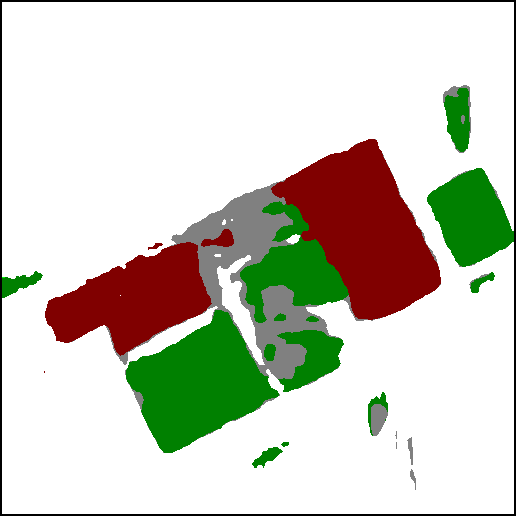} &
        \includegraphics[width=1.6cm]{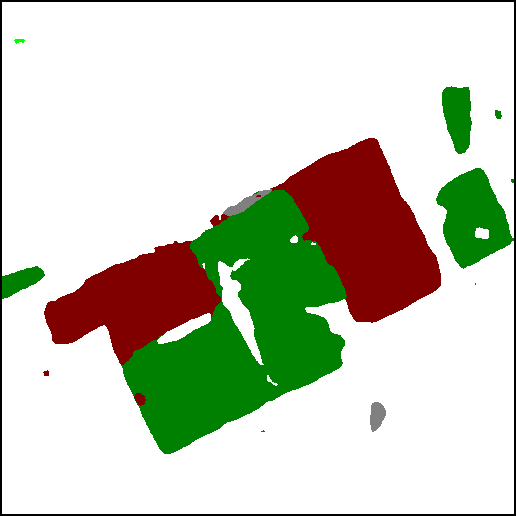} \\
        & Test image & GT & TED (w. $\mathcal{L}_{sc} \& \mathcal{L}_{psd}$) & SCanNet \\
    \end{tabular}
    \caption{Comparison of results obtained by the proposed technique w. and w/o. using the SCanFormer.} \label{Fig.Ablation_SCanFormer}
\end{figure}

3) \textbf{Effects of the SCanFormer}. Fig.\ref{Fig.Ablation_SCanFormer} compares the results obtained w. and w/o. using the proposed SCanFormer. While the detected changes are mostly the same, the SCanNet, leveraging the joint spatio-temporal information, can better discriminate the semantic categories. Specifically, in the results of the SCanNet there are less errors in the discrimination of \textit{ground} and \textit{building} (see Fig.\ref{Fig.Ablation_SCanFormer}(a)), as well as in the discrimination of \textit{low vegetation} and \textit{ground} (see Fig.\ref{Fig.Ablation_SCanFormer}(b)).

To conclude, this ablation study demonstrates that: i) the proposed TED framework and the learning scheme bring improvements to the detection of changes and the extraction of their semantic information; ii) the SCanFormer significantly improves the learning of temporal semantic information.

\subsection{Comparative Experiments}\label{sc5.compareSOTA}

To comprehensively evaluate the performance of the proposed method, we compare it with literature methods that are provided with accessible codes. The compared methods include i) the CD and SCD architectures proposed in \cite{daudt2018fully, daudt2019multitask}, ii) the hybrid CNN-RNN methods for SCD \cite{mou2018learning} and iii) other recent literature methods \cite{peng2021scdnet, ding2022bi}.

\begin{table*}[t]
    \centering
    \caption{Comparison with the SOTA methods for SCD.}
    \resizebox{1\linewidth}{!}{%
        \begin{tabular}{r|cccc|cccc}
        \toprule
            \multirow{2}*{Method} & \multicolumn{4}{c|}{SECOND} & \multicolumn{4}{c}{Landsat-SCD} \\
            \cline{2-9}
            & OA(\%) & mIoU(\%) & Sek(\%) & $F_{scd}$(\%) & OA(\%) & mIoU(\%) & Sek(\%) & $F_{scd}$(\%)\\
            \hline
            ResNet-GRU \cite{mou2018learning} & 85.09 & 60.64 & 8.99 & 45.89 & 90.55 & 74.16 & 26.51 & 71.87 \\
            ResNet-LSTM \cite{mou2018learning} & 86.91 & 67.27 & 16.14 & 57.05 & 93.36 & 80.88 & 40.06 & 80.36\\
            FC-Siam-conv. \cite{daudt2018fully} & 86.92 & 68.86 & 16.36 & 56.41 & 92.89 & 79.86 & 36.94 & 78.29 \\
            FC-Siam-diff \cite{daudt2018fully} & 86.86 & 68.96 & 16.25 & 56.20 & 91.95 & 76.44 & 30.23 & 73.97 \\
            HRSCD-str.2 \cite{daudt2019multitask} & 85.49 & 64.43 & 10.69 & 49.22 & 86.06 & 74.92 & 2.89 & 36.52 \\
            HRSCD-str.3 \cite{daudt2019multitask} & 84.62 & 66.33 & 11.97 & 51.62 & 91.10 & 78.33 & 31.43 & 73.17\\
            HRSCD-str.4 \cite{daudt2019multitask} & 86.62 & 71.15 & 18.80 & 58.21 & 91.27 & 79.10 & 32.29 & 73.34 \\
            SCDNet\cite{peng2021scdnet} & 87.46 & 70.97 & 19.73 & 60.01 & 94.94 & 85.23 & 50.05 & 85.00 \\
            SSCD-l\cite{ding2022bi} & 87.19 & 72.60 & 21.86 & 61.22 & 94.75 & 85.25 & 50.17 & 84.91 \\
            Bi-SRNet\cite{ding2022bi} & 87.84 & 73.41 & 23.22 & 62.61 & 94.91 & 85.53 & 51.01 & 85.35 \\
            \hline
            TED (proposed) & 87.39 & 72.79 & 22.17 & 61.56 & 95.89 & 88.49 & 58.69 & 88.22 \\
            SCanNet (proposed) & \textbf{87.86} & \textbf{73.42} & \textbf{23.94} & \textbf{63.66} & \textbf{96.26} & \textbf{88.96} & \textbf{60.53} & \textbf{89.27} \\
        \bottomrule
        \end{tabular} }\label{Table.CompareSOTA}
\end{table*}

1) \textbf{Quantitative Results}. We report the quantitative results obtained on the two benchmark datasets in Table.\ref{Table.CompareSOTA}. Among the compared methods, those SCD-targeted methods obtain higher accuracy metrics. Specifically, the ResNet-LSTM\cite{mou2018learning}, which contains temporal-modeling recurrent modules, has a relatively high $F_{scd}$. The HRSCD-str.4, which integrates CD and LCLU mapping, obtains higher mIoU and $F_{scd}$ on the SECOND. Among the literature methods, the SCDNet\cite{peng2021scdnet} that uses change-to-temporal attention operations obtains significantly higher accuracy. The Bi-SRNet\cite{ding2022bi} that enables interactions between the temporal semantic branches obtains the second-best and third-best accuracy on the SECOND and the Landsat-SCD dataset, respectively.

Without bells and whistles, plain TED outperforms most of the literature methods. Its advantages are more obvious in the Landsat-SCD dataset, which can be attributed to its spatial-preserving designs. The major difference between the TED and the HRSCD-str.4 is that the former reuses the large amount of semantic information presented in features from the temporal branches, therefore is more sensitive to the semantic changes. The proposed SCanNet enables deep and intrinsic modeling of the spatio-temporal dependencies in the SCD task. Thus it has significant advantages over the compared methods in all the metrics. Notably, its advantages over the second-best results in SeK are around 0.7\% in $SeK$ and around 1\% in $F_scd$. This demonstrates its superior capabilities in discriminating the semantic classes with a limited number of change samples.

\begin{figure*}
\centering
    {\includegraphics[width=14cm]{Pics/ST_colorbar.png}}\\
    \setlength{\tabcolsep}{1pt}
    \begin{tabular}{>{\centering\arraybackslash}m{0.6cm}>{\centering\arraybackslash}m{2.3cm}>{\centering\arraybackslash}m{2.3cm}>{\centering\arraybackslash}m{2.3cm}>{\centering\arraybackslash}m{2.3cm}>{\centering\arraybackslash}m{2.3cm}>{\centering\arraybackslash}m{2.3cm}>{\centering\arraybackslash}m{2.3cm}>{\centering\arraybackslash}m{2.3cm}}
        (a1) &
        \includegraphics[width=2.2cm]{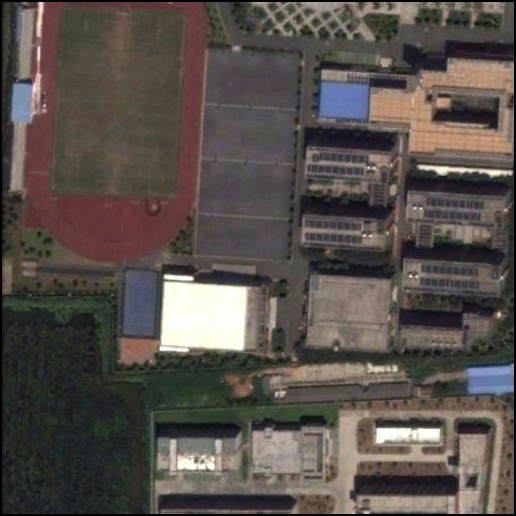} &
        \includegraphics[width=2.2cm]{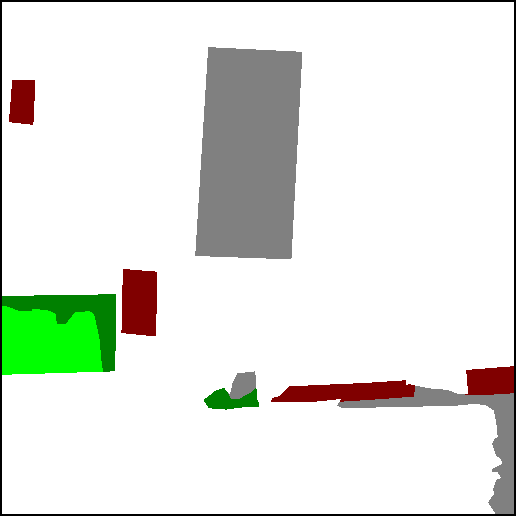} &
        \includegraphics[width=2.2cm]{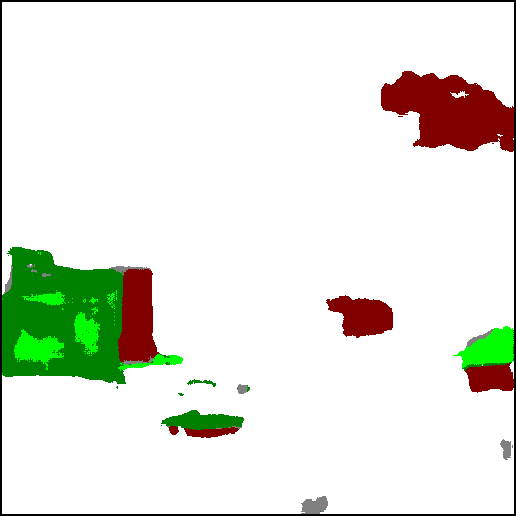} &
        \includegraphics[width=2.2cm]{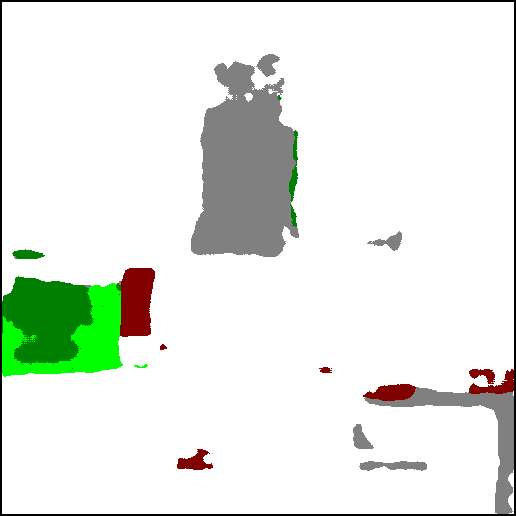} &
        \includegraphics[width=2.2cm]{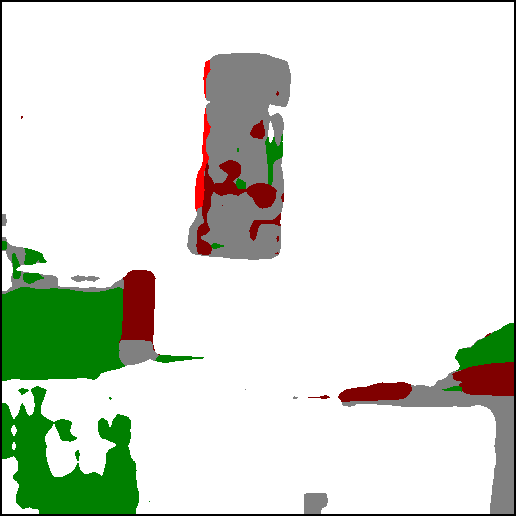} &
        \includegraphics[width=2.2cm]{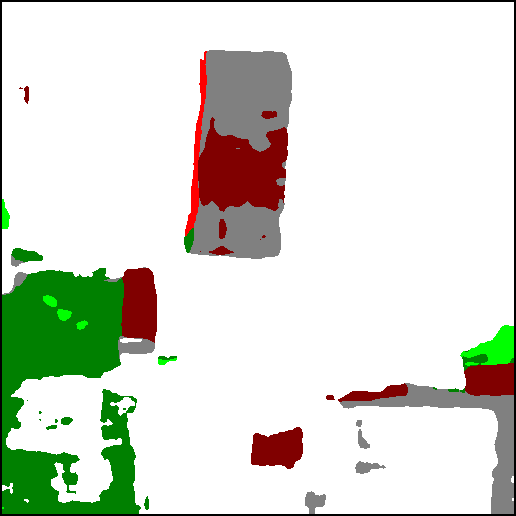} &
        \includegraphics[width=2.2cm]{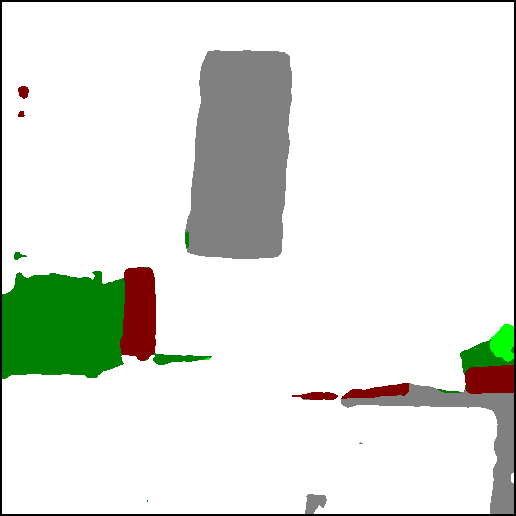} \\
        (a2) &
        \includegraphics[width=2.2cm]{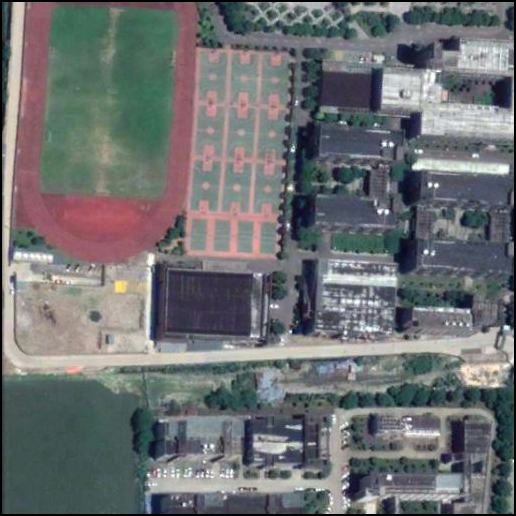} &
        \includegraphics[width=2.2cm]{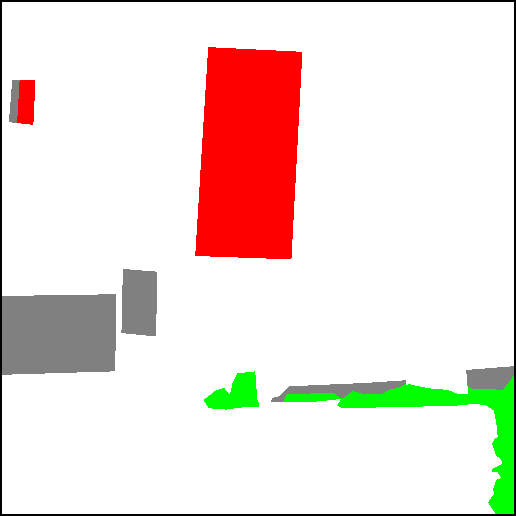} &
        \includegraphics[width=2.2cm]{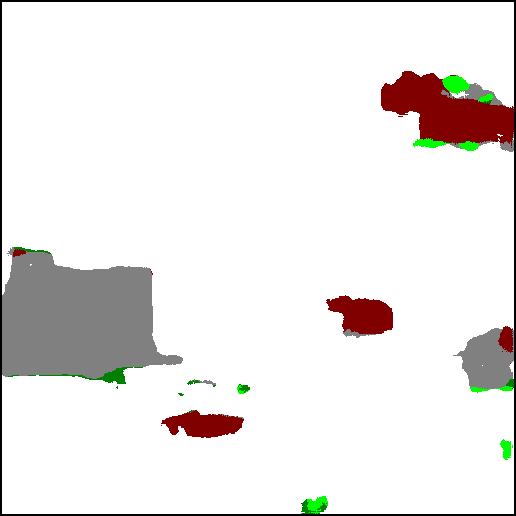} &
        \includegraphics[width=2.2cm]{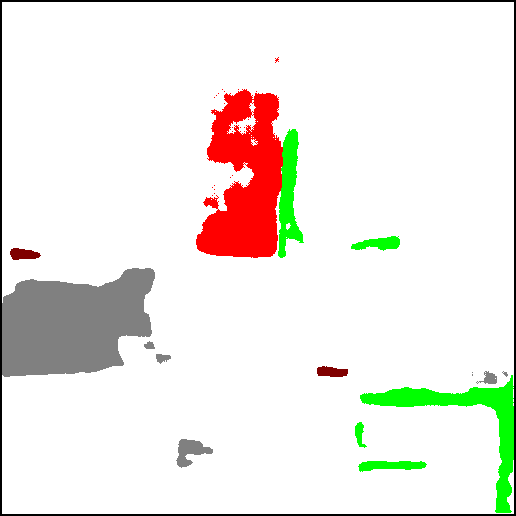} &
        \includegraphics[width=2.2cm]{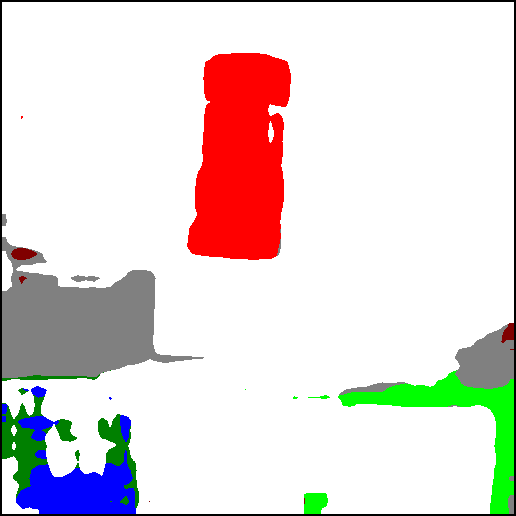} &
        \includegraphics[width=2.2cm]{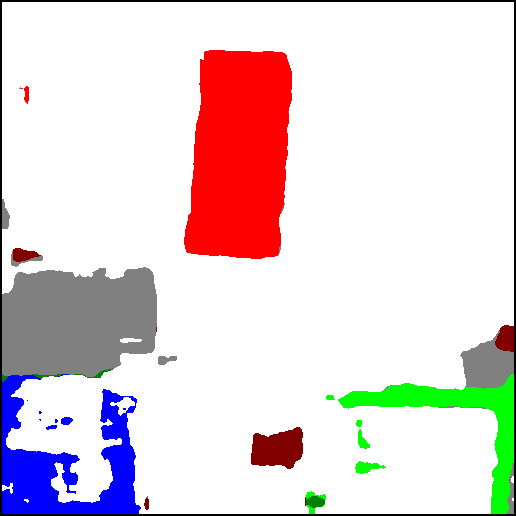} &
        \includegraphics[width=2.2cm]{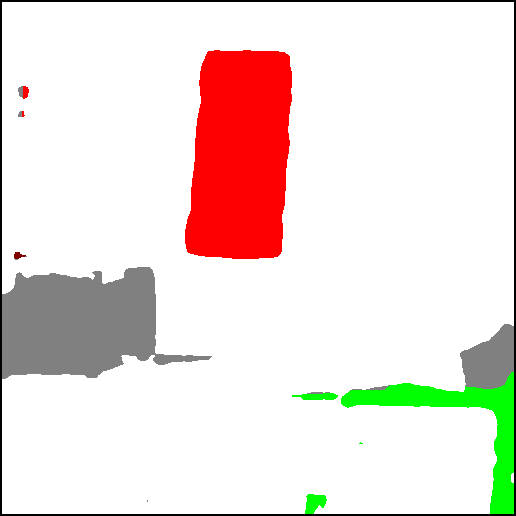} \\
        \hline\\
        (b1) &
        \includegraphics[width=2.2cm]{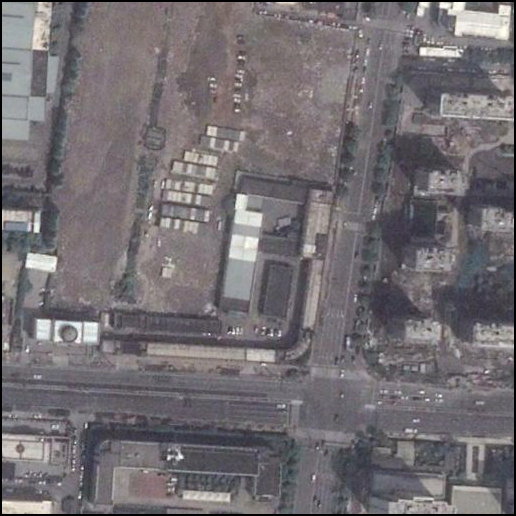} &
        \includegraphics[width=2.2cm]{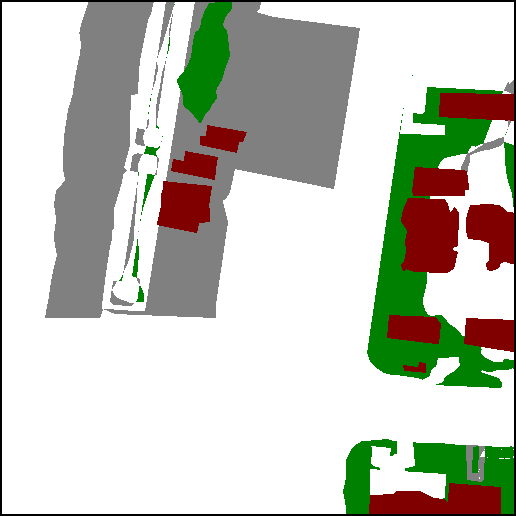} &
        \includegraphics[width=2.2cm]{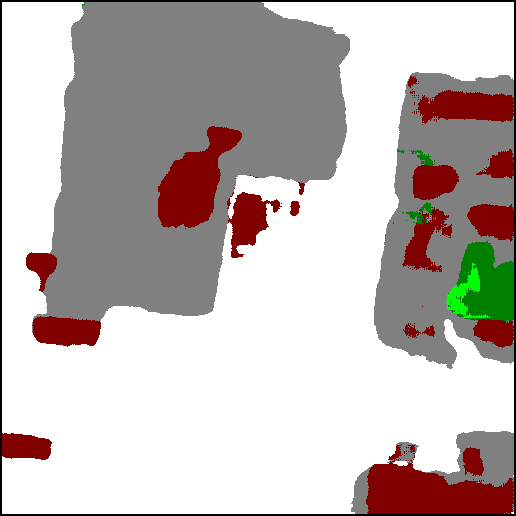} &
        \includegraphics[width=2.2cm]{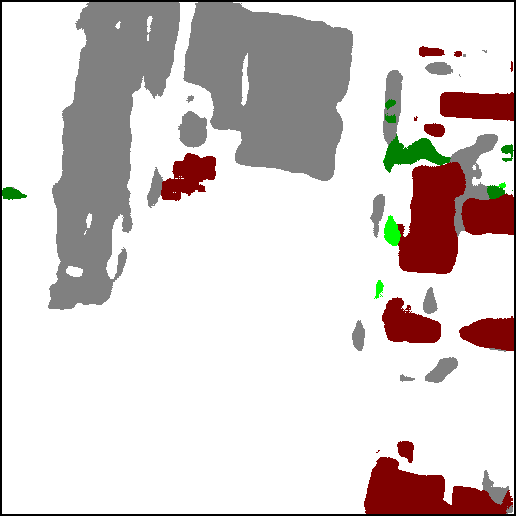} &
        \includegraphics[width=2.2cm]{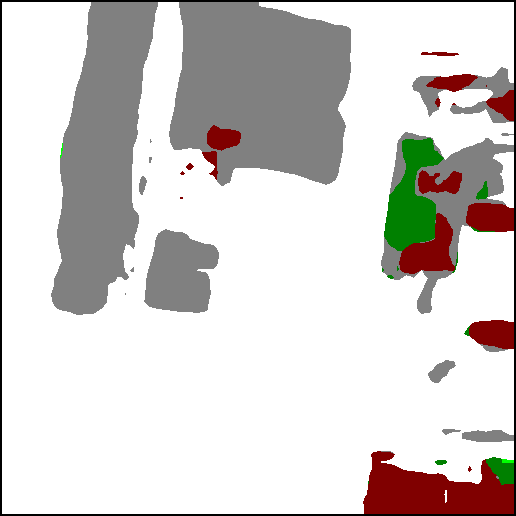} &
        \includegraphics[width=2.2cm]{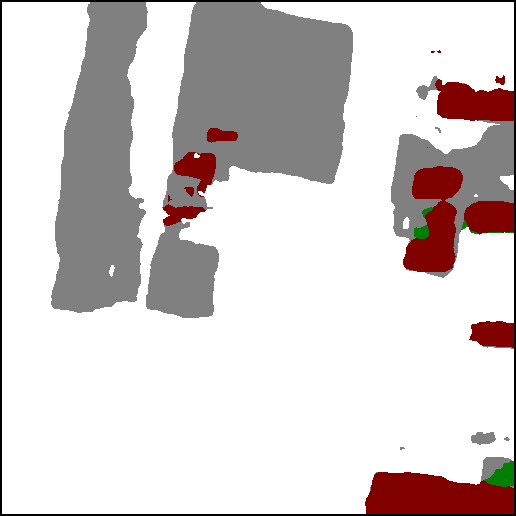} &
        \includegraphics[width=2.2cm]{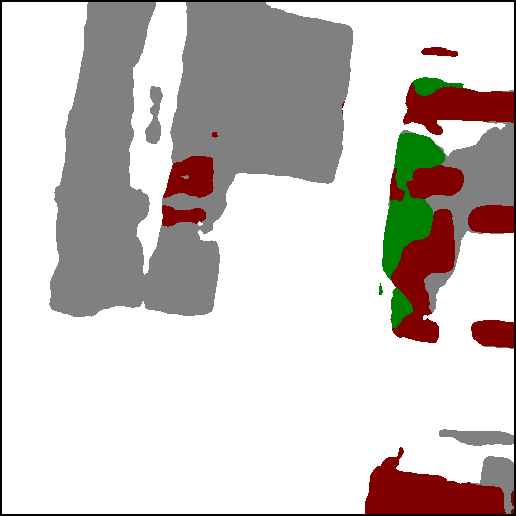} \\
        (b2) &
        \includegraphics[width=2.2cm]{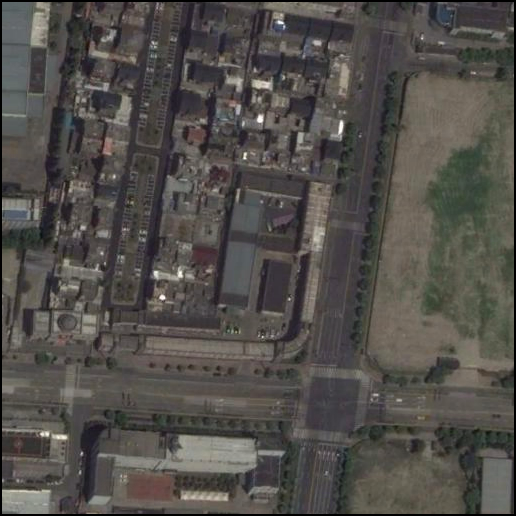} &
        \includegraphics[width=2.2cm]{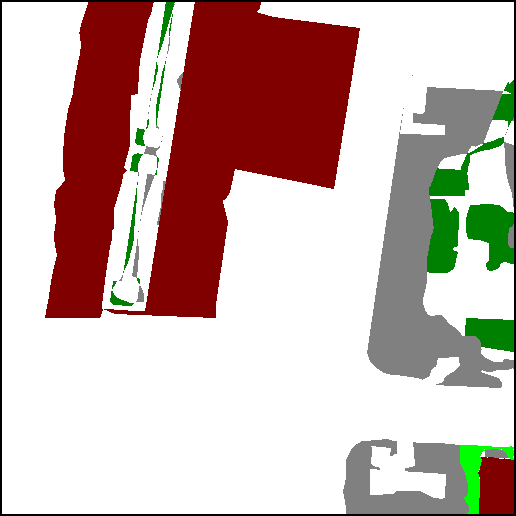} &
        \includegraphics[width=2.2cm]{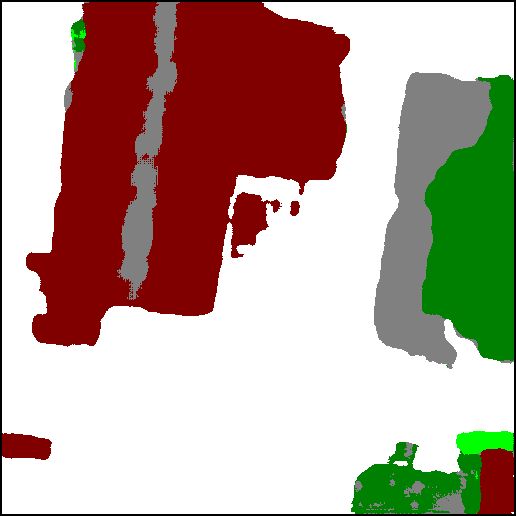} &
        \includegraphics[width=2.2cm]{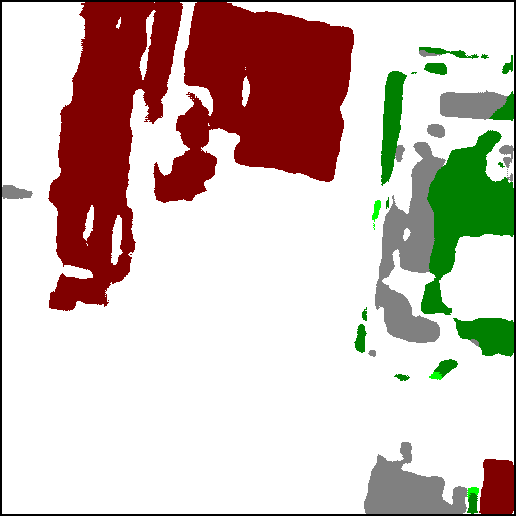} &
        \includegraphics[width=2.2cm]{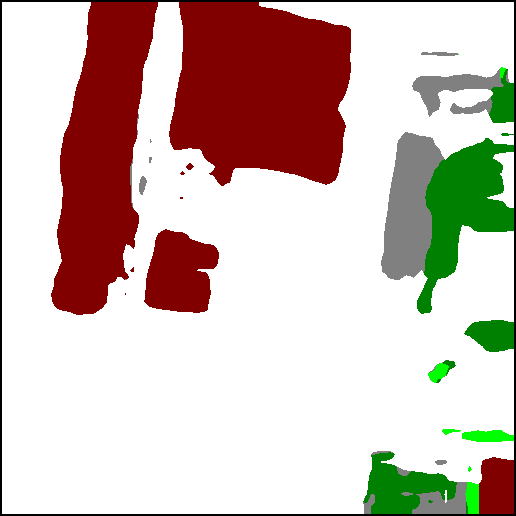} &
        \includegraphics[width=2.2cm]{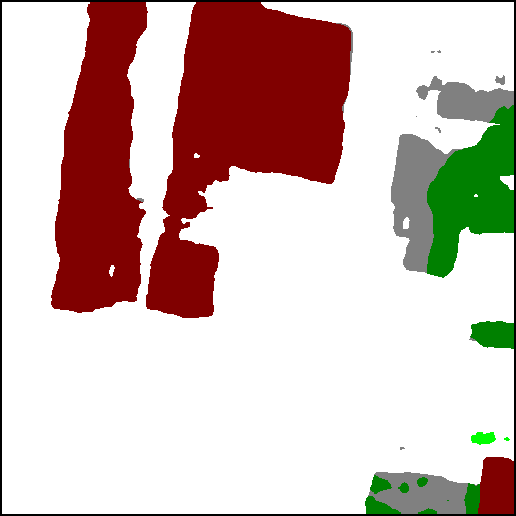} &
        \includegraphics[width=2.2cm]{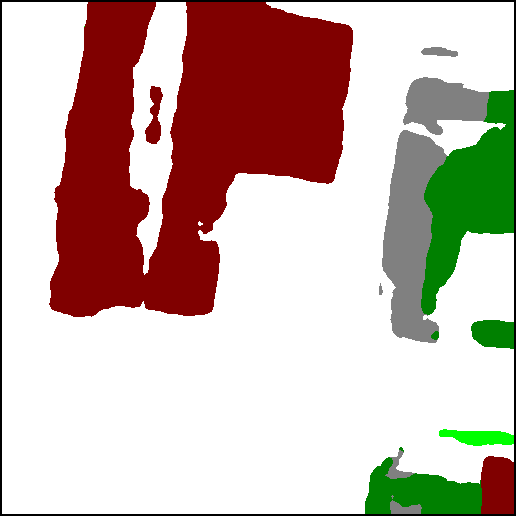} \\
        & Test image & GT & HRSCD-str.4 & SCDNet & BiSRNet & TED (proposed) & SCanNet (proposed)\\
        & & \includegraphics[width=10.0cm]{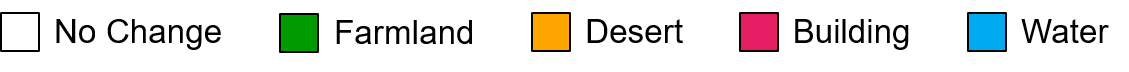}\\
        (c1) &
        \includegraphics[width=2.2cm]{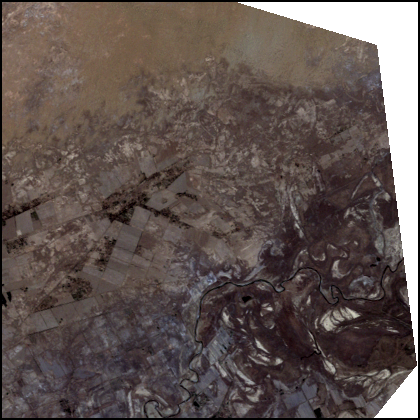} &
        \includegraphics[width=2.2cm]{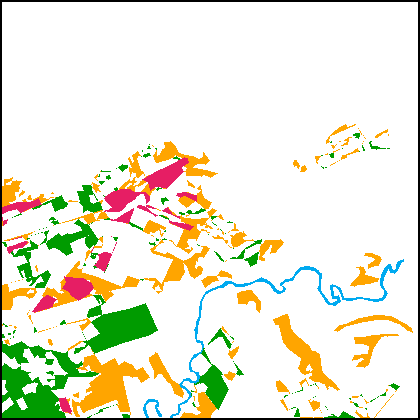} &
        \includegraphics[width=2.2cm]{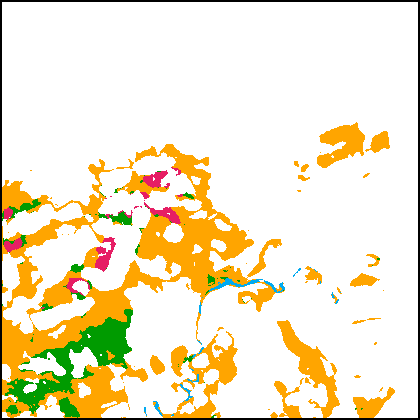} &
        \includegraphics[width=2.2cm]{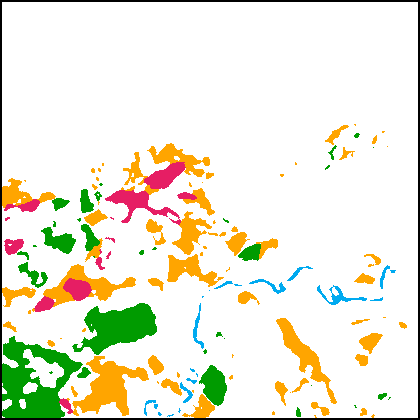} &
        \includegraphics[width=2.2cm]{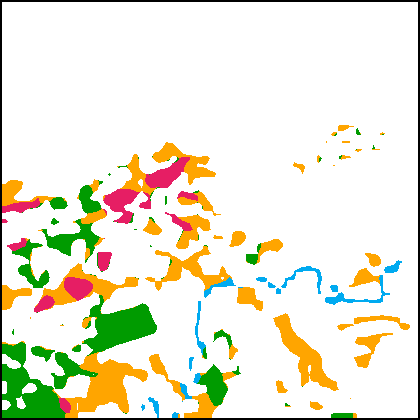} &
        \includegraphics[width=2.2cm]{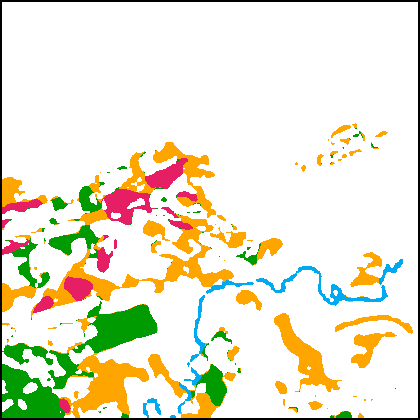} &
        \includegraphics[width=2.2cm]{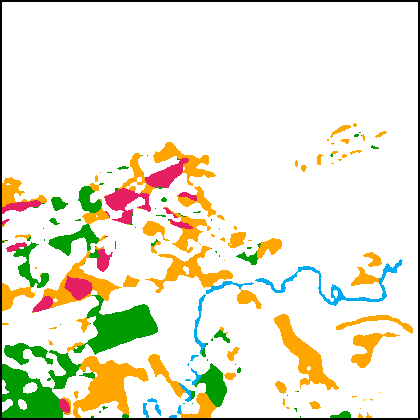} \\
        (c2) &
        \includegraphics[width=2.2cm]{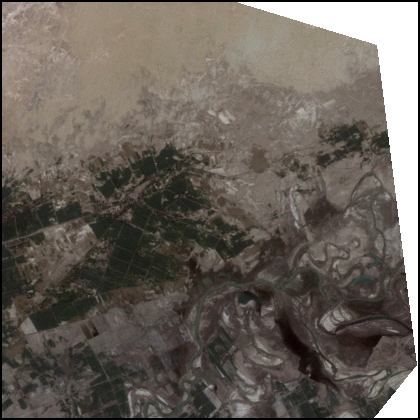} &
        \includegraphics[width=2.2cm]{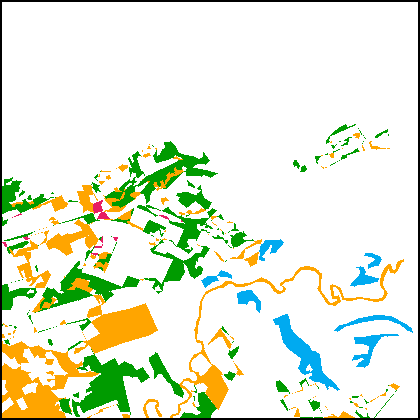} &
        \includegraphics[width=2.2cm]{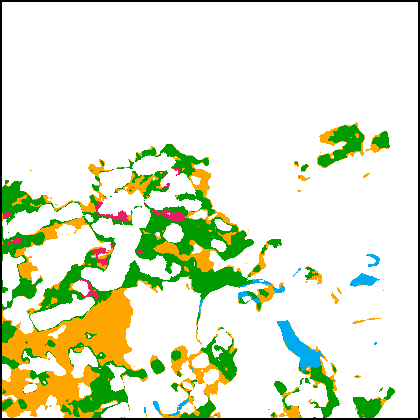} &
        \includegraphics[width=2.2cm]{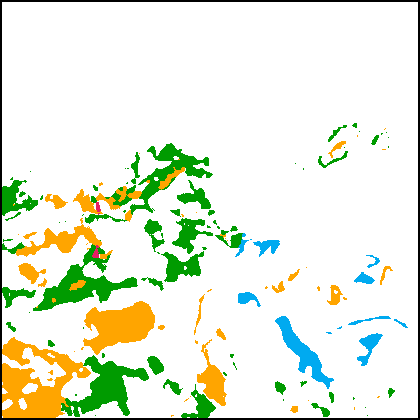} &
        \includegraphics[width=2.2cm]{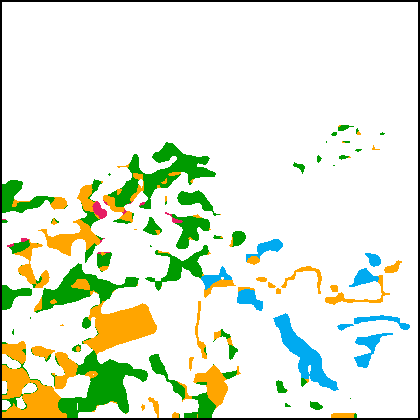} &
        \includegraphics[width=2.2cm]{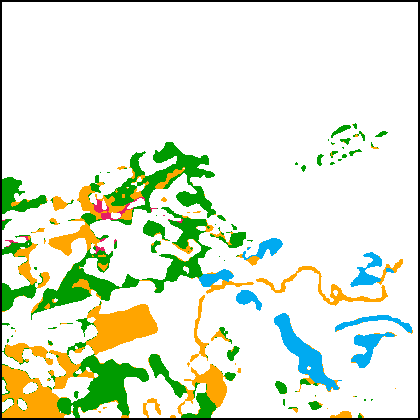} &
        \includegraphics[width=2.2cm]{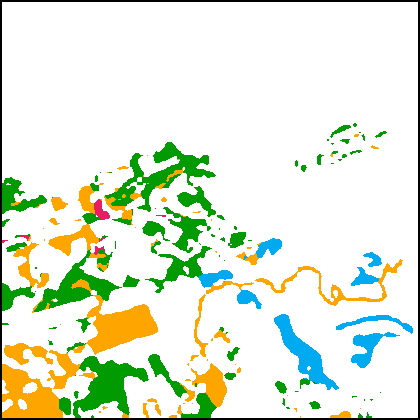} \\
        \hline\\
        (d1) &
        \includegraphics[width=2.2cm]{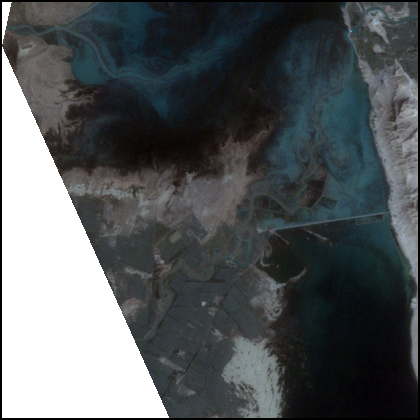} &
        \includegraphics[width=2.2cm]{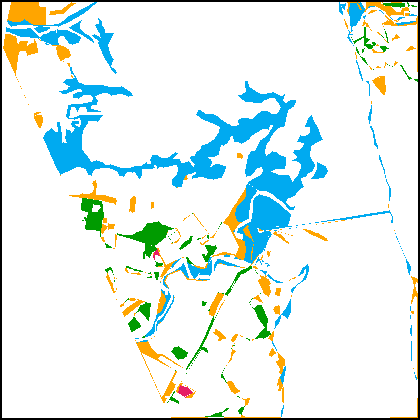} &
        \includegraphics[width=2.2cm]{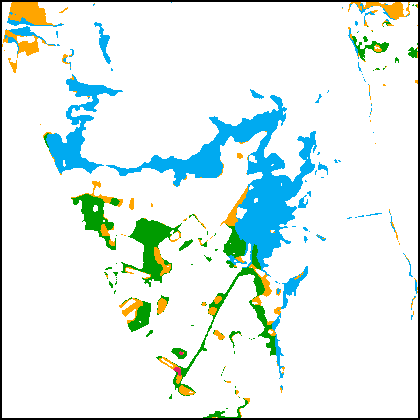} &
        \includegraphics[width=2.2cm]{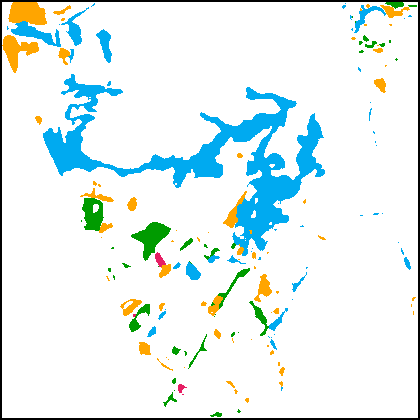} &
        \includegraphics[width=2.2cm]{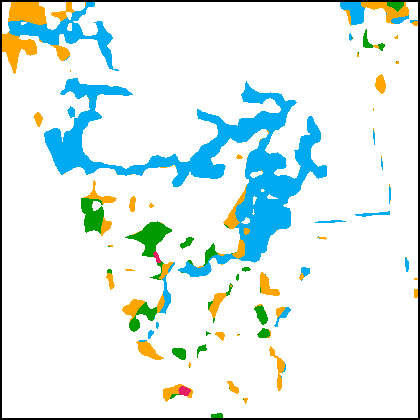} &
        \includegraphics[width=2.2cm]{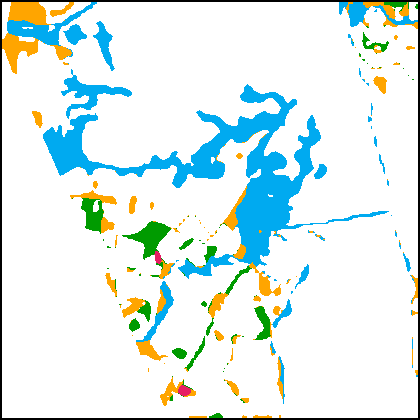} &
        \includegraphics[width=2.2cm]{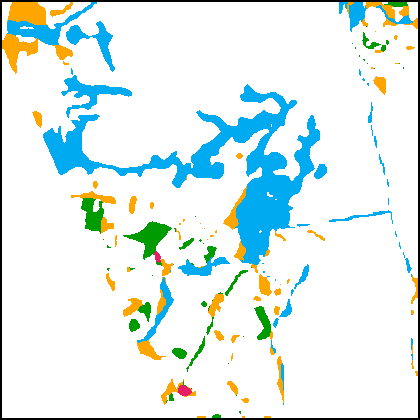} \\
        (d2) &
        \includegraphics[width=2.2cm]{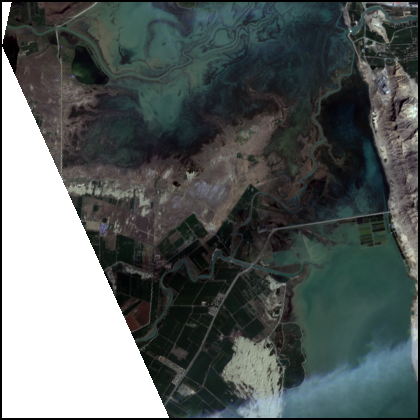} &
        \includegraphics[width=2.2cm]{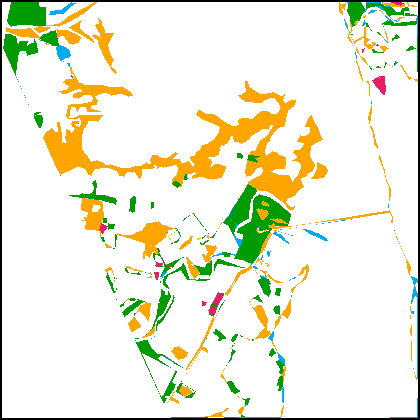} &
        \includegraphics[width=2.2cm]{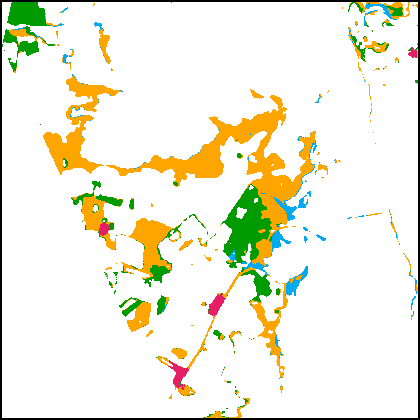} &
        \includegraphics[width=2.2cm]{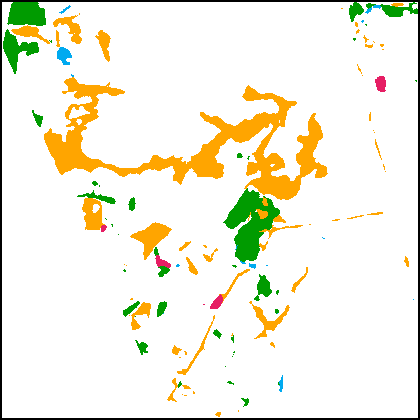} &
        \includegraphics[width=2.2cm]{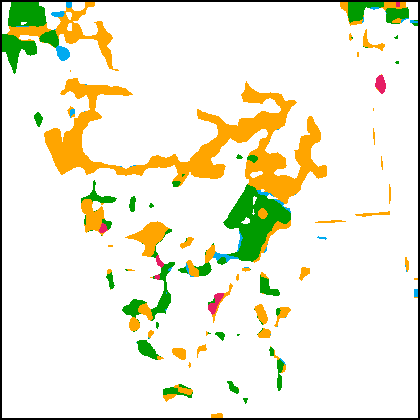} &
        \includegraphics[width=2.2cm]{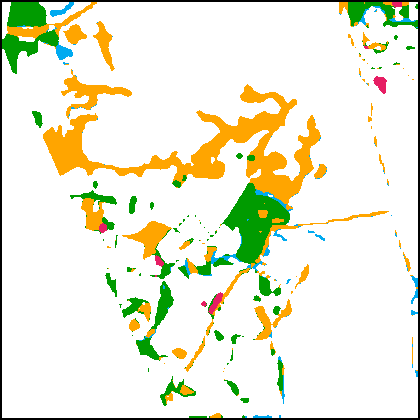} &
        \includegraphics[width=2.2cm]{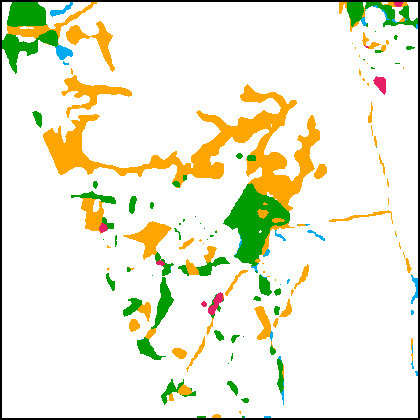} \\
        & Test image & GT & HRSCD-str.4 & SCDNet & BiSRNet & TED (proposed) & SCanNet (proposed)\\
    \end{tabular}
    \caption{Example of results provided by different methods in the comparative experiments. (a1)-(b2) results selected from the SECOND, (c1)-(d2) results selected from the Landsat-SCD dataset. Zoom in to find more details.} \label{Fig.SOTA} 
\end{figure*}

2) \textbf{Qualitative Results}. To visually compare the results, we present the segmentation maps obtained by different methods in Fig.\ref{Fig.SOTA}. The first 4 rows (Fig.\ref{Fig.SOTA}(a1)-(b2)) are results obtained on the SECOND. One can observe that the literature methods have difficulty in detecting the non-salient changes, e.g., the emergence of a playground in Fig.\ref{Fig.SOTA}(a) and the removal of small buildings in Fig.\ref{Fig.SOTA}(b). There are also discrepancies in the results. For example, in the SCD results of the HRSCD-str.4 and the SCDNet, there are areas that are segmented as \textit{building} on both of the bi-temporal segmentation maps, which is contradictory with respect to the represented change information. These issues are mostly addressed in the results of the proposed methods. With the guidance of semantic learning objectives, the detection of non-salient changes is improved and there are much fewer discrepancies in the bi-temporal results. The SCanNet further outperforms the plain TED in recognizing the critical areas, e.g., discrimination between \textit{low vegetation} and \textit{water} in Fig.\ref{Fig.SOTA}(a).

Fig.\ref{Fig.SOTA}(c1)-(d2) present SCD results obtained on the Landsat-SCD dataset. Since the GSD of this dataset is relatively low, it requires the SCD model to better preserve the spatial details. One can observe that the proposed methods based on the TED architecture can precisely capture fine-grained changes in LU types, such as the drying of a river in Fig.\ref{Fig.SOTA}(c), and the emergence of small \textit{farmlands} in Fig.\ref{Fig.SOTA}(d). The SCanNet shows advantages in discriminating the semantic categories of the small objects.

\begin{table*}
    \centering
    \caption{Major semantic changes detected by the different methods (sorted according to the confusion matrices in Fig.\ref{Fig.ChangeAnalysis}). The green texts indicate correct results, while the red texts indicate errors.}
    \resizebox{1\linewidth}{!}{%
        \begin{tabular}{r|cccccc}
        \toprule
            \multirow{2}*{Method} & \multicolumn{6}{c}{SECOND} \\
            \cline{2-7}
            & 1st & 2nd & 3rd & 4th & 5th & 6th \\
            \hline
            GT & ground$\rightarrow{}$building & low vegetation$\rightarrow{}$ground & ground$\rightarrow{}$low vegetation & low vegetation$\rightarrow{}$building & building$\rightarrow{}$ground & ground$\rightarrow{}$tree \\
            HRSCD-str.4 \cite{daudt2019multitask} & \textcolor{green}{ground$\rightarrow{}$building} & \textcolor{green}{low vegetation$\rightarrow{}$ground} & \textcolor{red}{low vegetation$\rightarrow{}$building} & \textcolor{red}{ground$\rightarrow{}$low vegetation} & \textcolor{green}{building$\rightarrow{}$ground} & \textcolor{red}{ground$\rightarrow{}$ground} \\
            SCDNet\cite{peng2021scdnet} & \textcolor{green}{ground$\rightarrow{}$building} & \textcolor{red}{low vegetation$\rightarrow{}$building} & \textcolor{red}{low vegetation$\rightarrow{}$ground} &  \textcolor{red}{building$\rightarrow{}$ground} & \textcolor{red}{ground$\rightarrow{}$low vegetation} & \textcolor{green}{ground$\rightarrow{}$tree} \\
            Bi-SRNet\cite{ding2022bi} & \textcolor{green}{ground$\rightarrow{}$building} & \textcolor{red}{low vegetation$\rightarrow{}$building} & \textcolor{red}{low vegetation$\rightarrow{}$ground} & \textcolor{red}{ground$\rightarrow{}$low vegetation} & \textcolor{green}{building$\rightarrow{}$ground} & \textcolor{green}{ground$\rightarrow{}$tree}\\
            SCanNet (proposed) & \textcolor{green}{ground$\rightarrow{}$building} & \textcolor{green}{low vegetation$\rightarrow{}$ground} & \textcolor{red}{low vegetation$\rightarrow{}$building} & \textcolor{red}{ground$\rightarrow{}$low vegetation} & \textcolor{green}{building$\rightarrow{}$ground} & \textcolor{green}{ground$\rightarrow{}$tree} \\
            \hline
            \multirow{2}*{Method} & \multicolumn{6}{c}{Landsat-SCD dataset} \\
            \cline{2-7}
            & 1st & 2nd & 3rd & 4th & 5th & 6th \\
            \hline
            GT & desert$\rightarrow{}$farmland & desert$\rightarrow{}$water & farmland$\rightarrow{}$desert & water$\rightarrow{}$desert & farmland$\rightarrow{}$building & desert$\rightarrow{}$building \\
            HRSCD-str.4 \cite{daudt2019multitask} & \textcolor{green}{desert$\rightarrow{}$farmland} & \textcolor{green}{desert$\rightarrow{}$water} & \textcolor{red}{desert$\rightarrow{}$desert} & \textcolor{red}{farmland$\rightarrow{}$desert} & \textcolor{red}{farmland$\rightarrow{}$farmland} & \textcolor{red}{water$\rightarrow{}$desert} \\
            SCDNet\cite{peng2021scdnet} & \textcolor{green}{desert$\rightarrow{}$farmland} & \textcolor{green}{desert$\rightarrow{}$water} & \textcolor{green}{farmland$\rightarrow{}$desert} & \textcolor{green}{water$\rightarrow{}$desert} & \textcolor{green}{farmland$\rightarrow{}$building} & \textcolor{green}{desert$\rightarrow{}$building}  \\
            Bi-SRNet\cite{ding2022bi} & \textcolor{green}{desert$\rightarrow{}$farmland} & \textcolor{green}{desert$\rightarrow{}$water} & \textcolor{green}{farmland$\rightarrow{}$desert} & \textcolor{green}{water$\rightarrow{}$desert} & \textcolor{green}{farmland$\rightarrow{}$building} & \textcolor{green}{desert$\rightarrow{}$building} \\
            SCanNet (proposed) & \textcolor{green}{desert$\rightarrow{}$farmland} & \textcolor{green}{desert$\rightarrow{}$water} & \textcolor{green}{farmland$\rightarrow{}$desert} & \textcolor{green}{water$\rightarrow{}$desert} & \textcolor{green}{farmland$\rightarrow{}$building} & \textcolor{green}{desert$\rightarrow{}$building} \\
        \bottomrule
        \end{tabular} }\label{Table.ChangeAnalysis}
\end{table*}

\begin{figure*}[!ht]
\centering
        \subcaptionbox{GT}
        {\includegraphics[height=3.4cm]{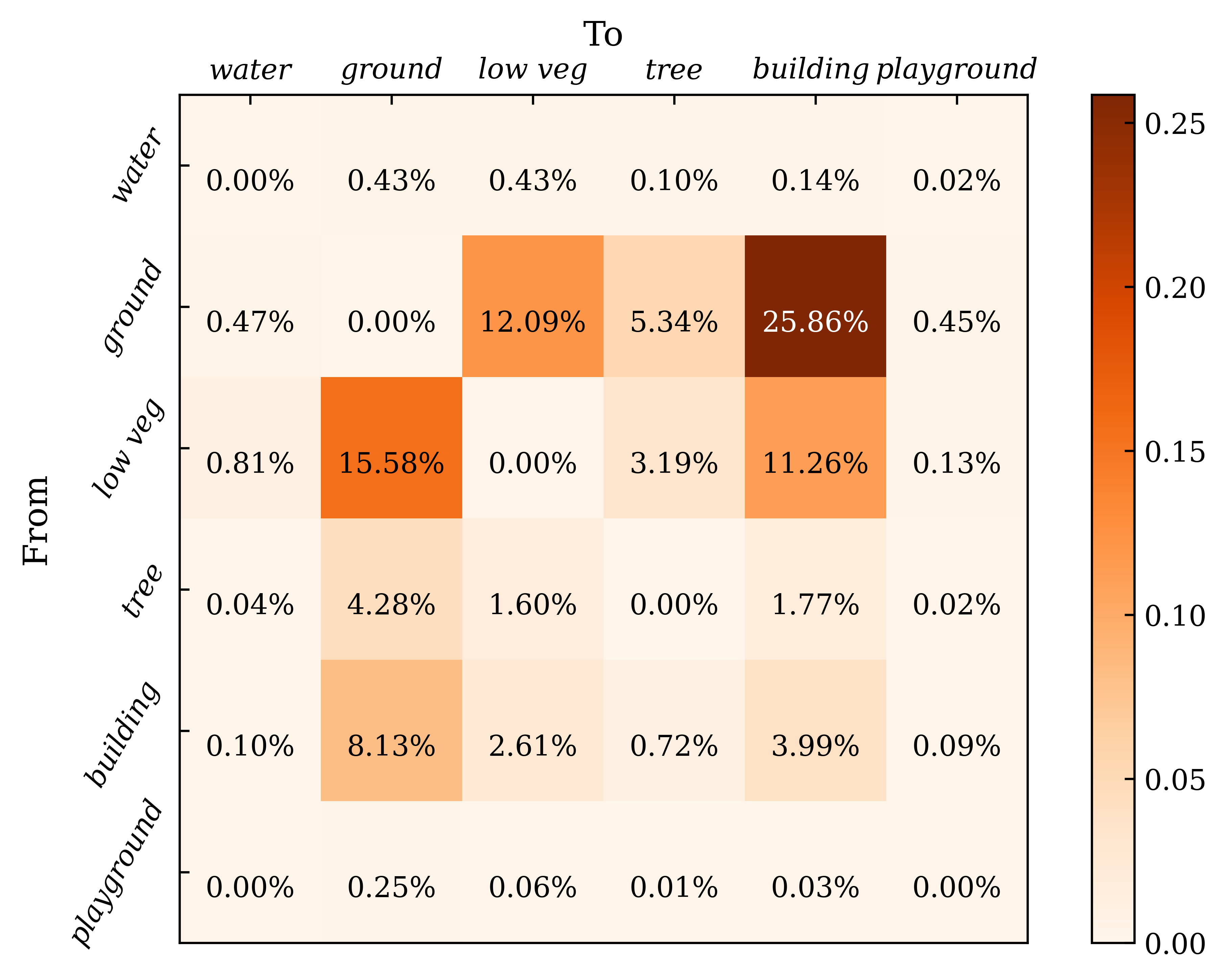}}
        \subcaptionbox{HRSCD-str.4}
        {\includegraphics[height=3.4cm]{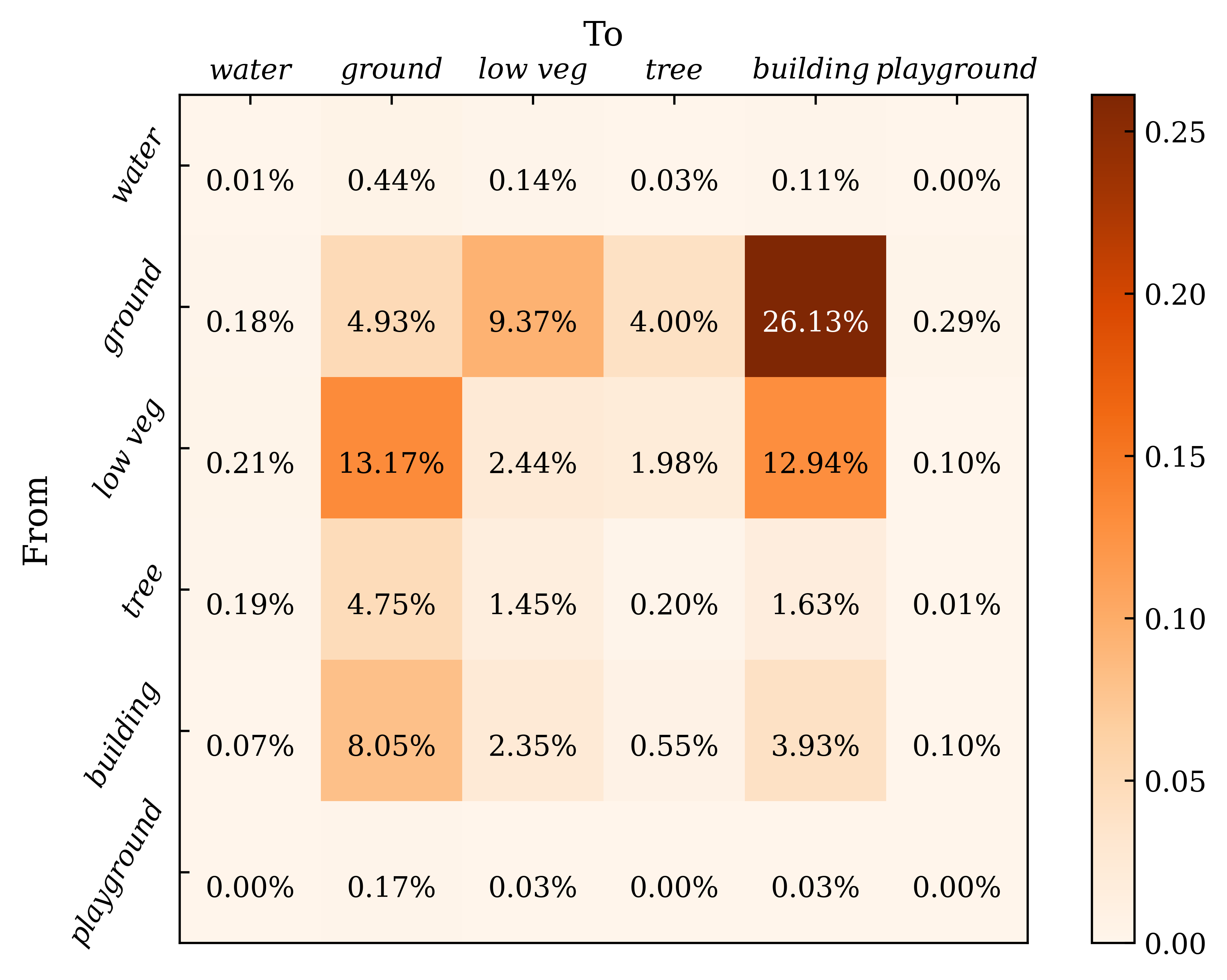}}
        \subcaptionbox{SCDNet}
        {\includegraphics[height=3.4cm]{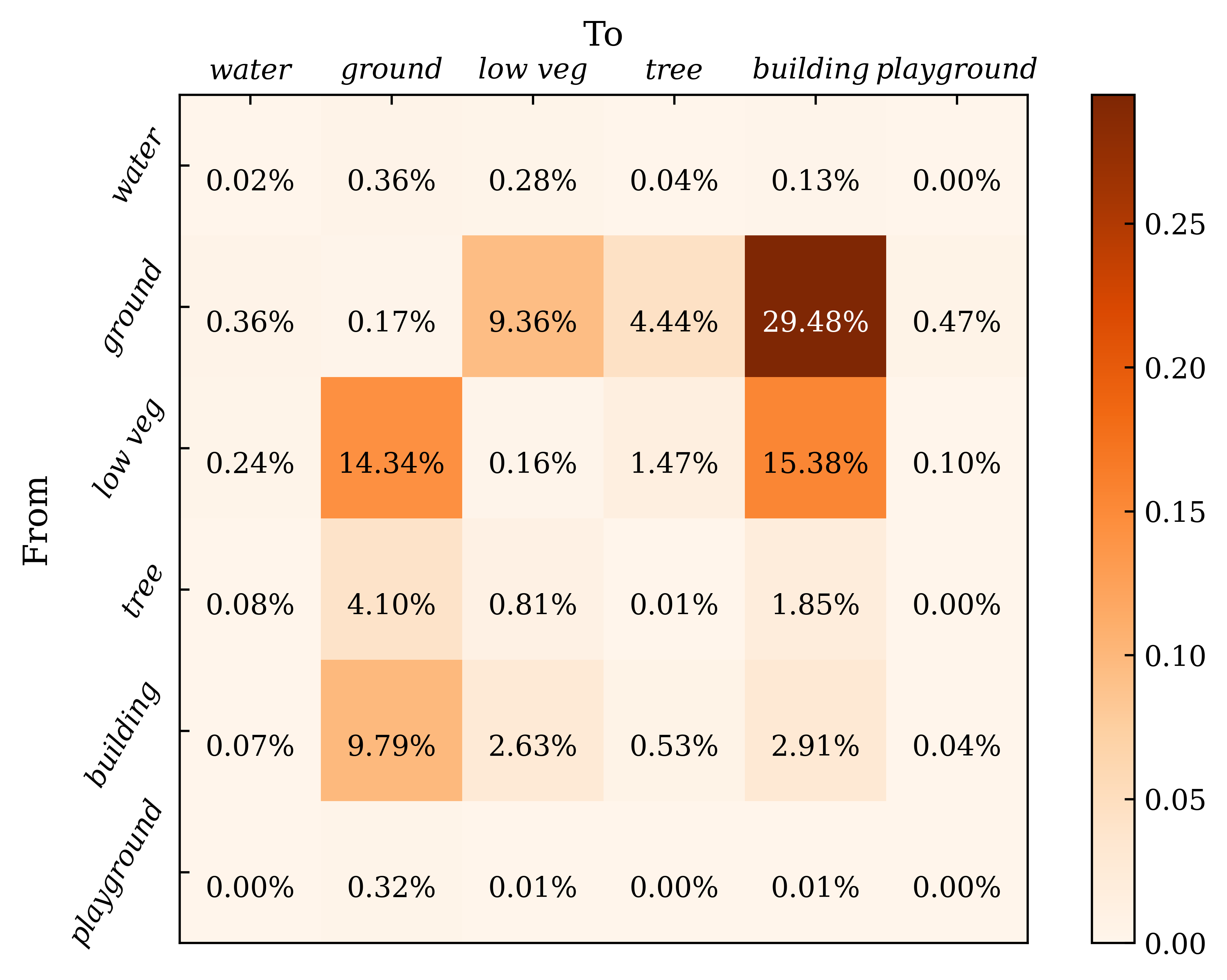}}
        \subcaptionbox{SCanNet}
        {\includegraphics[height=3.4cm]{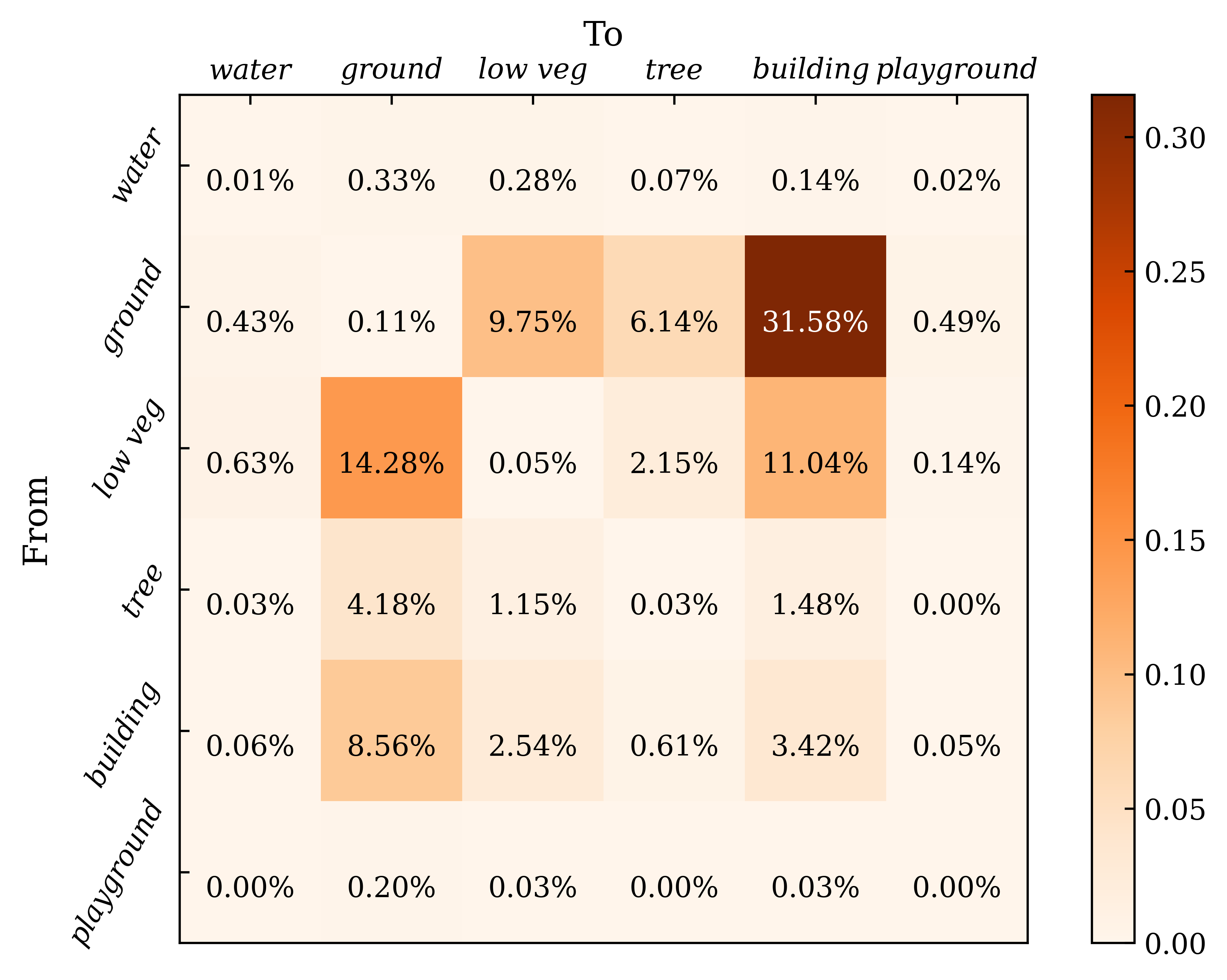}}\\
        \subcaptionbox{GT}
        {\includegraphics[height=3.4cm]{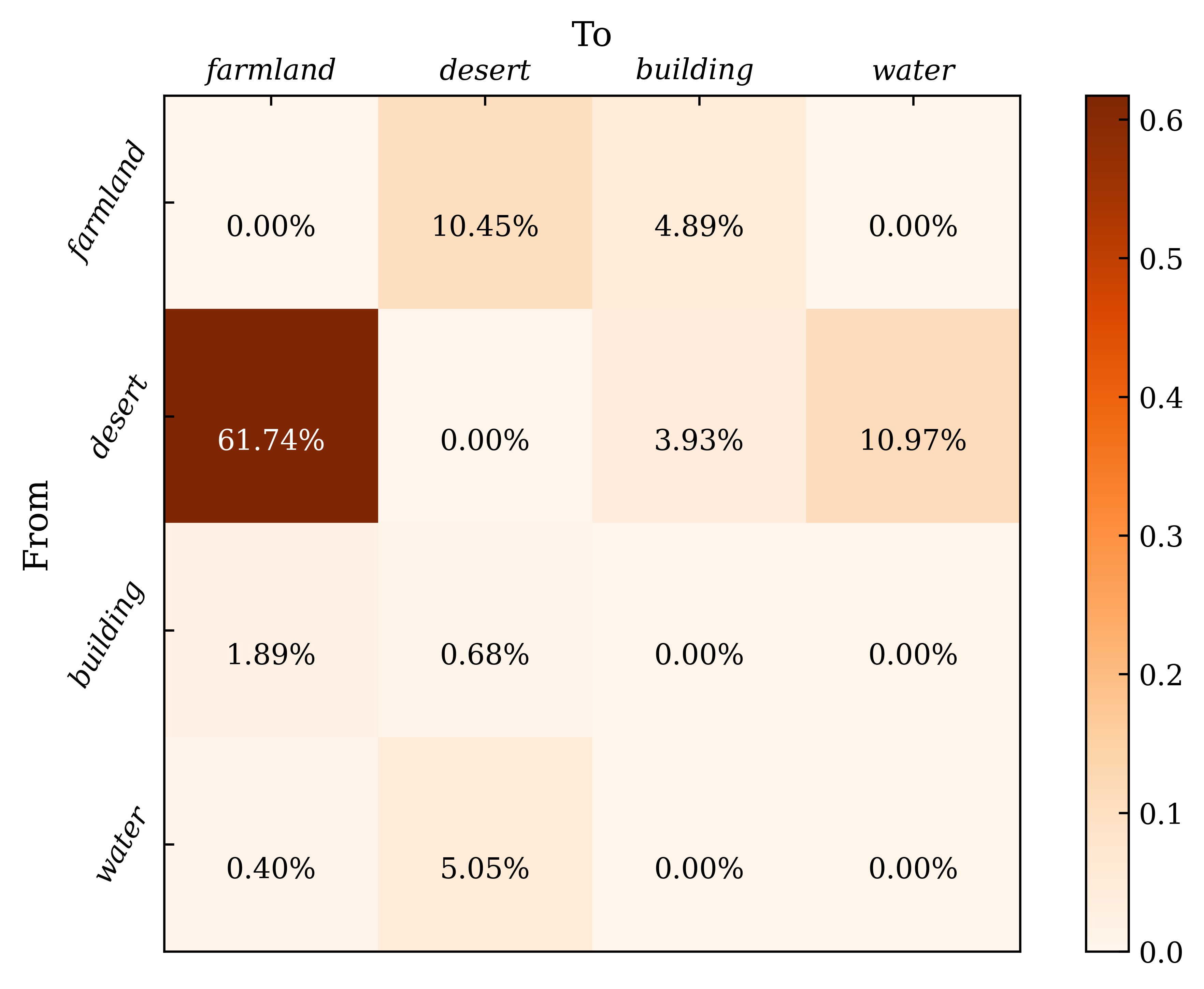}}
        \subcaptionbox{HRSCD-str.4}
        {\includegraphics[height=3.4cm]{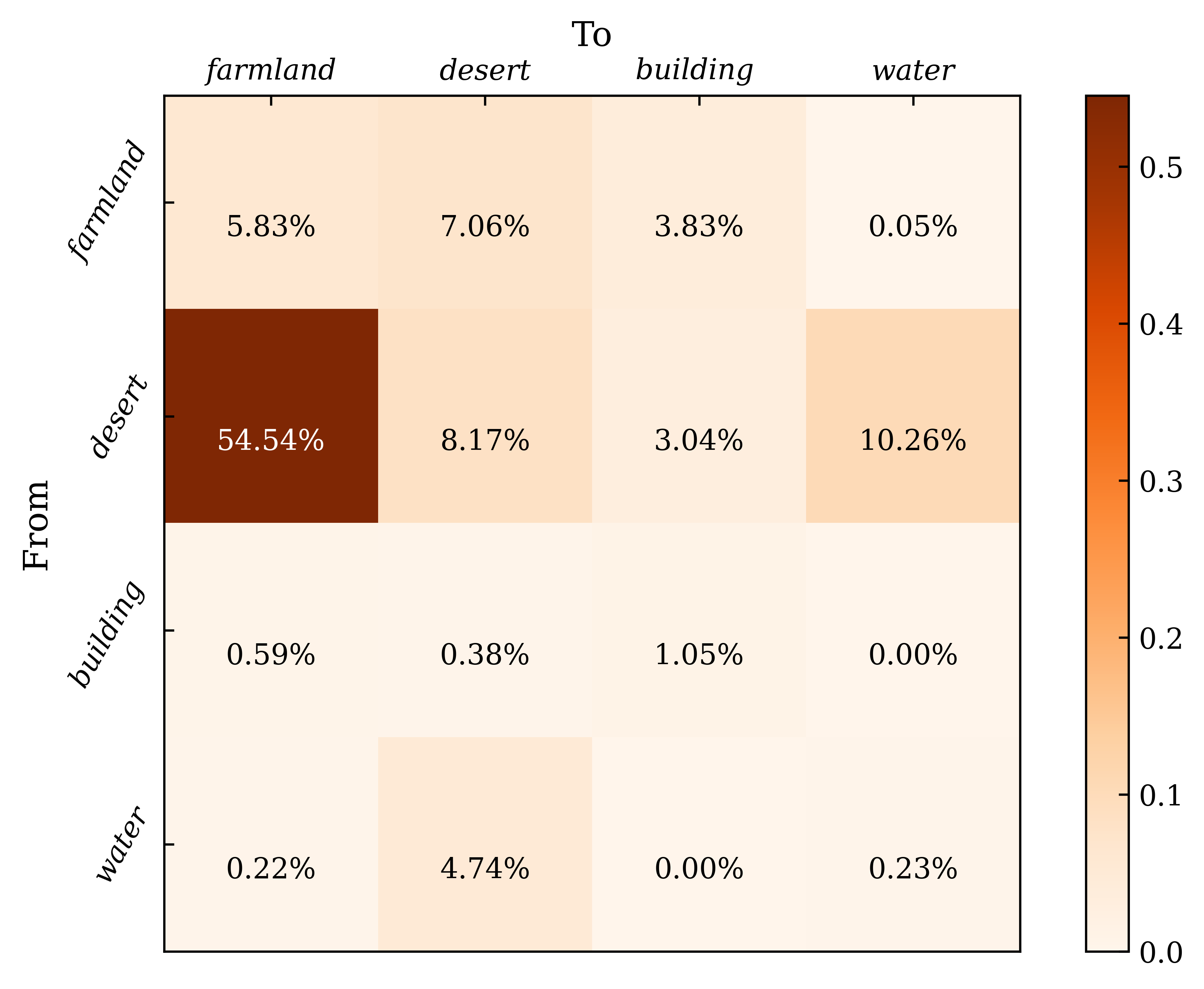}}
        \subcaptionbox{SCDNet}
        {\includegraphics[height=3.4cm]{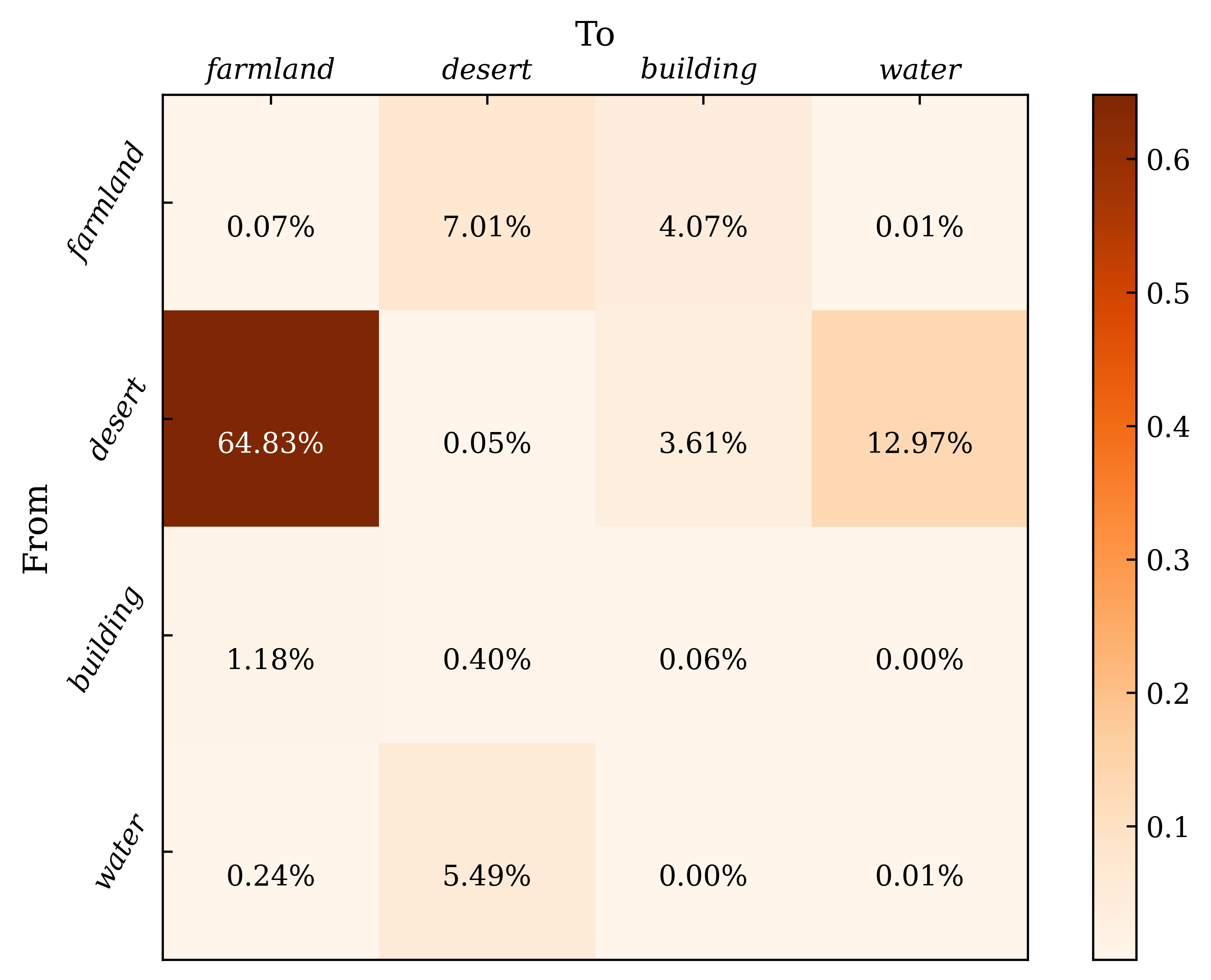}}
        \subcaptionbox{SCanNet}
        {\includegraphics[height=3.4cm]{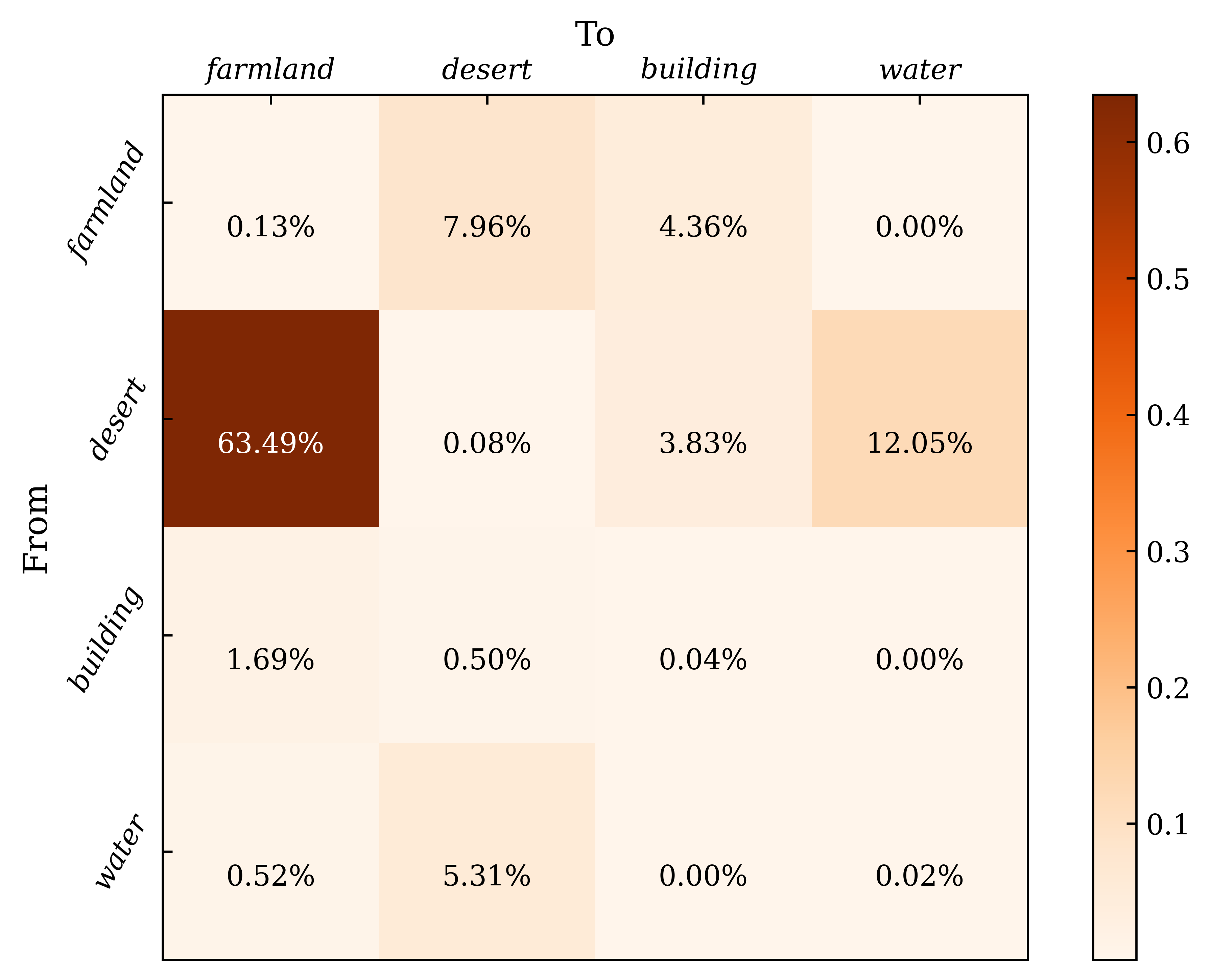}}
    \caption{The 'from-to' semantic change matrixes derived from the SCD results. (a)-(d) are derived from the SCD results in the SECOND, (e)-(h) are derived from the SCD results in the Landsat-SCD.}\label{Fig.ChangeAnalysis}
\end{figure*}

3) \textbf{Change Analysis}. In Fig.\ref{Fig.ChangeAnalysis} we present the confusion matrices obtained on the bi-temporal SCD results, which intuitively present the 'from-to' change types in the observed regions. The GT labels in the SECOND (see Fig.\ref{Fig.ChangeAnalysis}(a)) indicate that the major semantic changes include \textit{ground$\rightarrow{}$building}, \textit{low vegetation$\rightarrow{}$ground}, \textit{ground$\rightarrow{}$low vegetation} and \textit{low vegetation$\rightarrow{}$building}, which account for 25.86\%, 15.58\%, 12.09\% and 11.26\% of the change proportions, respectively. This change pattern is generally consistent with the results of the SCanNet. To compare the semantic changes detected by the different methods, Table.\ref{Table.ChangeAnalysis} sorts the top-6 major changes derived from the change confusion matrices. One can observe that in the results of the SCanNet, there are fewer errors in terms of the sequence of the major semantic changes. Meanwhile, in the Landsat-SCD dataset, the major changes include \textit{desert$\rightarrow{}$farmland}, \textit{desert$\rightarrow{}$water}, \textit{farmland$\rightarrow{}$desert} and \textit{water$\rightarrow{}$desert}, accounting for 61.74\%, 10.97\%, 10.45\% and 5.05\% of the total changes, respectively. This dataset is less challenging, thus most of the methods successfully detected these changes.

It is worth noting that the results of the HRSCD.str4 (in the SECOND and in the Landsat) and the SCDNet (in the SECOND) show some discrepancies (i.e., false changes such as \textit{ground$\rightarrow{}$ground} and \textit{farmland$\rightarrow{}$farmland}). Meanwhile, in the results of the SCDNet, the false changes are limited to only a small proportion. There are a total of 0.2\% false changes in the SECOND (excluding the \textit{building$\rightarrow{}$building} changes which are affected by wrong annotations), and 0.26\% in the Landsat-SCD dataset. In another word, most of the discrepancies in the results are eliminated.
\section{Conclusions}\label{sc6}

SCD is a meaningful and challenging task in RS applications. It requires not only exploitation of the semantic and change features, but also comprehensive learning of the spatio-temporal dependencies, which are not fully considered in literature methods. We investigate to overcome this limitation in two-fold. First, we propose a hybrid CNN-transformer architecture for the SCD. The CNN branches effectively exploit the spatial context, while the SCanFormer is designed at the head of the network to jointly model the spatio-temporal dependencies. Second, we propose a semantic learning scheme that formulates the prior constraints contained in SCD. It contains a semantic consistency objective to boost the consistency in segmentation results, and a pseudo-learning objective to supervise the SE in \textit{no-change} areas. Thus, in a word, the SCanFormer provides capabilities for joint spatio-temporal analysis, while the learning scheme drives the SCanNet to model the intrinsic change-semantic correlations in SCD.

Extensive experiments have been conducted to evaluate the performance of the proposed method. The results obtained on the SCD benchmark dataset indicate that the proposed SCanNet has achieved accuracy improvements over the SOTA methods (with a lead of over 1\% in $F_{scd}$). One of the remaining challenges is to discriminate the rare semantic changes with very few samples, which is left for future studies.


\bibliographystyle{IEEEtran}
\bibliography{refs}

\end{document}